\documentclass[lettersize,journal]{IEEEtran}
\usepackage{amsmath,amsfonts}
\usepackage{algorithmicx}
\usepackage{algorithm}
\usepackage{array}
\usepackage[caption=false,font=footnotesize,labelfont=rm,textfont=rm]{subfig}
\usepackage{textcomp}
\usepackage{stfloats}
\usepackage{url}
\usepackage{verbatim}
\usepackage{graphicx}
\usepackage{cite}
\usepackage{ulem}
\usepackage{hyperref}
\usepackage{amsthm}
\usepackage{booktabs}
\usepackage[switch]{lineno}
\usepackage{bm}
\usepackage{amssymb}
\usepackage{algpseudocode}

\begin{document}

\title{MCoCo: Multi-level Consistency Collaborative Multi-view Clustering}

\author{Yiyang Zhou$^1$, Qinghai Zheng$^2$, Wenbiao Yan$^1$, Yifei Wang$^1$, Pengcheng Shi$^1$, {Jihua Zhu$^{1, *}$}

$^1$School of Software Engineering, Xi’an Jiaotong University, Xi’an 710049, China

$^2$College of Computer and Data Science, Fuzhou University, Fuzhou 350108, China
\thanks{*Corresponding author: Jihua Zhu (email: zhujh@xjtu.edu.cn).}}
\markboth{}%
% \markboth{Journal of \LaTeX\ Class Files,~Vol.~14, No.~8, August~2021}
{Shell \MakeLowercase{\textit{et al.}}: Multi-level consistency collaborative multi-view
clustering}


\maketitle

\begin{abstract}
Multi-view clustering can explore consistent information from different views to guide clustering.
Most existing works focus on pursuing shallow consistency in the feature space and integrating the information of multiple views into a unified representation for clustering.
These methods did not fully consider and explore the consistency in the semantic space.
To address this issue, we proposed a novel Multi-level Consistency Collaborative learning framework (MCoCo) for multi-view clustering.
Specifically, MCoCo jointly learns cluster assignments of multiple views in feature space and aligns semantic labels of different views in semantic space by contrastive learning.
Further, we designed a multi-level consistency collaboration strategy, which utilizes the consistent information of semantic space as a self-supervised signal to collaborate with the cluster assignments in feature space.
Thus, different levels of spaces collaborate with each other while achieving their own consistency goals, which makes MCoCo fully mine the consistent information of different views without fusion.
Compared with state-of-the-art methods, extensive experiments demonstrate the effectiveness and superiority of our method. Our code is released on \href{https://github.com/YiyangZhou/MCoCo}{https://github.com/YiyangZhou/MCoCo}.

\end{abstract}

\begin{IEEEkeywords}
Multi-view clustering, Consistency collaborative, Semantic consensus information.
\end{IEEEkeywords}

\section{Introduction}
\IEEEPARstart{M}{lti-view} data are collected by different collectors and feature extractors, and different views show heterogeneity.
Compared with the traditional single-view data, it is informative and can provide a more comprehensive description of objects \cite{geng2021uncertainty,zhang2018generalized,zhang2019ae2,zheng2020feature, jia2021multi}.
Thanks to these advantages, multi-view clustering (MVC) has attracted more and more attention in recent years.
Existing MVC methods can be roughly divided into traditional methods and deep methods.

Traditional methods can be further divided into three sub-categories: (1) Subspace-based clustering methods \cite{zhang2018generalized,yin2018multiview,xie2020robust,guo2021rank,kang2020large, chen2021low, lan2021generalized}, where a shared low-dimensional representation that integrates multi-view information and a similarity matrix is mined for clustering.
(2) Clustering method based on non-negative matrix decomposition \cite{liu2013multi,wang2018multiview,yang2020uniform}, which decomposes each view into a low-rank matrix for clustering.
(3) Graph-based clustering method \cite{nie2017self,wang2019gmc,peng2019comic, zheng2022graph, wong2019clustering, wang2022towards}, mining graph structure information to guide multi-view clustering.

\begin{figure}[t]
\centering
\subfloat[]{
    \includegraphics[width=0.45\linewidth]{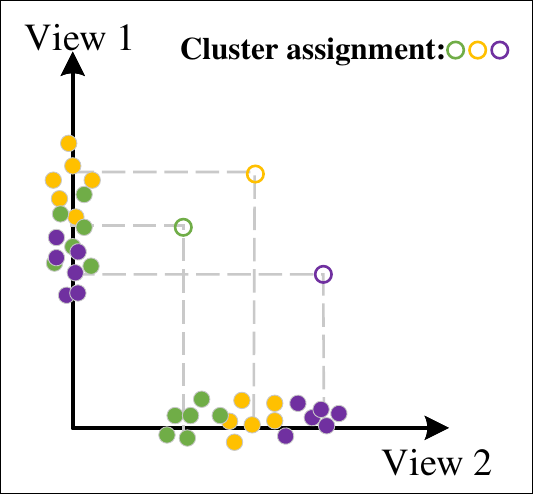}
    }\hfill
\subfloat[]{
    \includegraphics[width=0.45\linewidth]{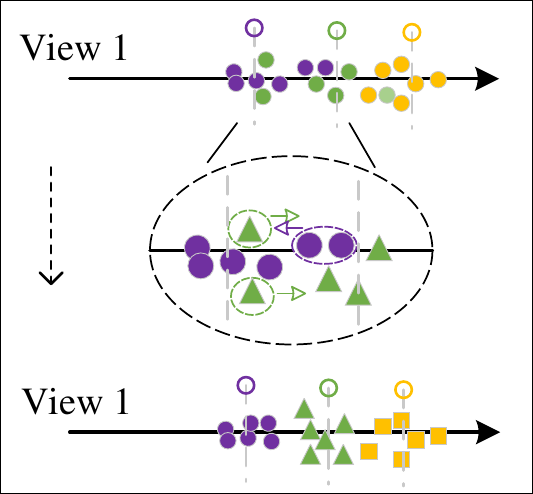}
    }
\caption{ An illustrative example of our motivation.
(a) Showing the mapping between two views. (b) Taking view 1 as an example, it shows the collaborative learning process of semantic labels and cluster assignments. If there are three categories in the dataset, different categories are represented by different colors, and same shapes represent samples with similar semantic information.}
\label{fg1}
\end{figure}

The deep neural network has shown excellent performance in many fields in recent years \cite{mildenhall2021nerf,xu2019learning, tao2017deep}.
In order to utilize the ability of the deep network to capture nonlinear features and deal with clustering tasks of large-scale data \cite{li2019deep,xu2021multi}, many MVC methods based on the deep network have appeared recently \cite{zhang2019ae2,zhang2018generalized,xu2022multi,geng2021uncertainty,lin2021completer,zheng2021collaborative}.
Most of them focus on integrating the information of multiple views into a comprehensive representation and pursuing the consistency of different views only in feature space, ignoring that the fusion of multiple views' features may cause some views with fuzzy clustering structure to interfere with the performance of the final representation, which may decline the performance of the model with the increase of the number of multi-view data views.

Aiming at these problems, some research on the non-fusion MVC method appeared \cite{xu2021multi,xu2022multi,xu2022self,xu2021deep}.
In order to obtain consistent cluster assignments of different views in feature space for clustering, most of them align the cluster assignments of different views to the cluster assignment of the common feature.
Compared with fusing multiple views into a complete representation, the non-fusion model can avoid the negative impact of the view with a fuzzy cluster structure.
As shown in Figure \ref{fg1}(a), the two separated views have clear cluster assignment mapping in the global space, so view 2 with a clear cluster structure can guide the cluster separation of view 1 by using the cluster assignments collaboration of different views.
The existing non-fusion MVC methods basically focus on the alignment of cluster assignments in feature space.
As shown at the top of Figure \ref{fg1}(b), although view 1 can be separated into three clusters through collaborative training of other views' cluster assignments, it is difficult to separate some overlapping areas in low-dimensional feature space because the computational essence of cluster assignment is mapping the distance between the low-dimensional features of views and their respective cluster centers into pseudo-labels \cite{xie2016unsupervised,cheng2021multi,li2019deep}.

\begin{figure*}[ht] 
\centering
\includegraphics[width=0.9\textwidth]{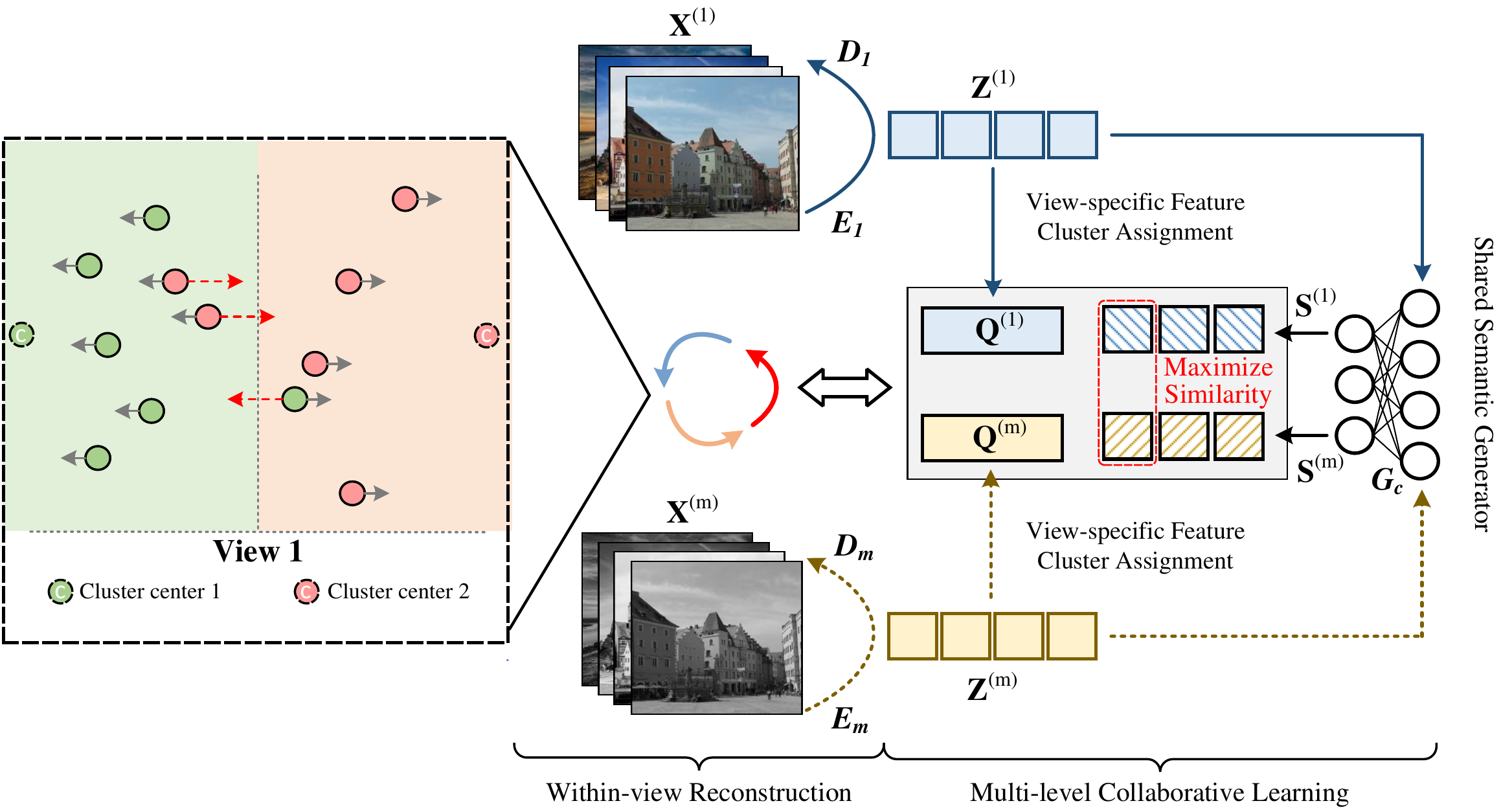}
\caption{Overview of MCoCo. Given $m$ views, each view $\mathbf{X}^{(m)}$ is mapped to its feature space $\mathbf{Z}^{(m)}$ by encoder $\bm{E}_m$ and decoder $\bm{D}_m$. $\mathbf{Z}^{(m)}$ is utilized to construct the cluster assignment $\mathbf{Q}^{(m)}$ by Eq.\ref{ep8}. Based on a shared semantic generator $\bm{G}_c$, $\mathbf{Z}^{(m)}$ is mapped into the semantic space $\mathbf{S}^{(m)}$, where the semantic labels $\{\mathbf{S}^{(i)}\}_{i=1}^m$ tend to be consistent by contrastive learning. Then, $\mathbf{Q}^{(m)}$ and $\mathbf{S}^{(m)}$ collaborate with each other to mine the multi-level consistent information. The left part illustrates the influence of one view under multi-level collaborative learning: the gray arrow indicates that the cluster assignments $\{\mathbf{Q}^{(i)}\}_{i=1}^{m}$ are jointly learned to make the sample closer to the cluster center, and the red arrow indicates the process of multi-level collaboration to correct the misassigned samples, where the samples with similar semantic information are forced to be close to each other.
}
\label{fg2} 
\end{figure*}

To solve the problems in the above discussion, we present a novel Multi-level Consistency Collaborative learning framework (MCoCo).
As illustrated in Figure \ref{fg2}, MCoCo consists of
two modules, namely, within-view reconstruction and multi-level collaborative learning.
We consider that the data of different views describe the same object, which makes it reasonable to exploit the consensus in the semantic space.
Further, in the multi-level collaborative learning module, a well-designed multi-level consistency collaboration strategy uses consistent semantic information to help the clustering assignments in feature space in a self-supervised manner.
As shown in Figure \ref{fg1}(b), suppose that there are three categories in the dataset.
Different categories are represented by different colors, and the same shape indicates that the samples with similar semantic information.
In feature space, we jointly learn the cluster assignments of each view, while in semantic space, the semantic labels of different views tend to be similar through contrastive learning.
Then, as shown in the middle of Figure \ref{fg1}(b), under the collaboration of the semantic labels the misassigned samples can be corrected.
The consistent information of different levels of spaces collaborate with each other, which enables MCoCo to learn discriminating clustering assignments and mine the multi-level consistent information in multiple views.

The main contributions of the proposed method can be summarized as follows:

\begin{itemize}
\item We introduce a novel Multi-level Consistency Collaborative learning framework (MCoCo) for multi-view clustering, which can mine multi-level consistent information to guide clustering under the collaboration of consistent information in different levels of spaces.

\item We propose a brand-new multi-level consistency collaboration strategy that allows MCoCo to achieve the respective consistency goals of feature space and semantic space while realizing multi-level space collaboration.

\item Extensive experiments are conducted on diverse benchmark datasets, and experimental results demonstrate its state-of-the-art clustering performance.
\end{itemize}

\section{Related Work}
\subsubsection{Deep Single-view Clustering}
The emergence of deep neural networks has led to rapid development in various fields of computer science, and deep networks have been used by researchers to handle clustering tasks. The most representative work in deep single-view clustering is \cite{xie2016unsupervised}, which is an end-to-end learning method that can automatically learn feature representations and cluster assignments for data. It maps data to a low-dimensional feature space and iteratively optimizes a clustering objective to achieve simultaneous learning of feature representations and cluster assignments.
Subsequently, many variant versions emerged in the continuation of this work \cite{guo2017improved, ghasedi2017deep}.
\cite{mukherjee2019clustergan} is a new mechanism proposed for clustering using GANs. This is achieved by sampling from a mixture of one-hot encoded variables and continuous latent variables, and training an inverse network which projects the data to the clustering latent space.
These methods have achieved impressive progress in the field of deep single-view clustering. However, they can only handle individual views and cannot effectively leverage the rich information provided by multi-view data to enhance clustering performance.

\subsubsection{Multi-view Clustering}
Multi-view clustering is a challenging and important branch of multi-view learning research that aims to explore the rich semantic information in multiple views and use this information to guide clustering in low-dimensional space \cite{zhao2017multi, xu2013survey}.
For multi-view clustering, the key is to explore the consistency information among multiple views. CCA-based methods \cite{wang2015deep, hotelling1992relations, akaho2006kernel, andrew2013deep} aim to maximize the canonical correlation between different views to uncover the consistent information among them. For two views, the paradigm of CCA-based methods can be summarized as follows:
\begin{equation}
\label{ep0}
    \min_{\beta _{1}, \beta _{2}} -corr(  \bm{f}_{1}( \mathbf{X}^{(1) }  ; \beta _{1}) , \bm{f}_{ 2}( \mathbf{X}^{(2) }  ; \beta _{2})) + \lambda reg( \beta _{1},\beta _{2}),
\end{equation}
where $\bm{f}_{1}(\cdot; \beta_{1})$ and $\bm{f}_{2}(\cdot; \beta_{2})$ are two embedding strategies with parameters $\beta _{1}$ and $\beta _{2}$. $corr(\cdot)$ and $reg(\cdot)$ indicate the canonical correlation function and the regularization term respectively.
In \cite{andrew2013deep} $\bm{f}_{1}(\cdot; \beta_{1})$ and $\bm{f}_{2}(\cdot; \beta_{2})$ are both deep neural networks. As for \cite{wang2015deep} $\bm{f}_{1}(\cdot; \beta_{1})$ and $\bm{f}_{2}(\cdot; \beta_{2})$ are utilized to learn bottleneck representations of two autoencoders.
In addition, some CCA-based methods \cite{shao2016deep, shen2015unified, chen2018canonical} also consider utilizing graph information to guide clustering.

More, \cite{zheng2023comprehensive} integrates low-dimensional embedding representations and applies low-rank tensor constraints on the subspace representations of multiple views to construct a comprehensive feature representation, incorporating rich information across the views.
\cite{huang2021deep} is capable of generating a unified multi-view spectral representation through the introduction of an orthogonal constraint and reformulation strategy that utilizes Cholesky decomposition during the learning process. This approach enables the model to effectively capture and exploit information across multiple views.
\cite{wan2021multill} relies on the information bottleneck principle to integrate shared representation among different views and view-specific representation of each view, promoting a comprehensive representation of multi-views and flexibly balancing the complementarity and consistency among multiple views.
\cite{hu2021akm} is a novel multi-view clustering method that uses m sub-cluster centers to reveal the sub-cluster structure in multi-view data, thereby improving clustering performance. To fully utilize complementary information between different views, it uses a multi-view combination weights strategy to automatically assign weights and properly fuse information from different views to obtain an optimally shared bipartite graph.
\cite{zheng2021collaborative} and \cite{zheng2022graph} explored high-order correlations and graph information in multiple views in a shared representation space, respectively, and achieved equally encouraging results.

Most existing methods focus on integrating multiple views into a shared low-dimensional space for clustering, while this paper differs from the majority of methods by utilizing consistent information across different views at multiple levels to guide learning of a discriminative clustering assignment. The non-fusion strategy can avoid negative impacts from views with ambiguous clustering assignments during fusion.

\subsubsection{Contrastive Learning}
Contrastive learning \cite{chen2020simple, xie2021detco} is one of most effective unsupervised representation learning paradigm that aims to minimize the spatial distance or maximize the similarity between positive pairs, while maximizing the spatial distance between them and their corresponding negative pairs.
In recent years, contrastive learning has made remarkable progress in representation learning and computer vision, such as \cite{niu2022spice} and \cite{van2020scan}. With the utilization of deep networks in multi-view clustering, contrastive learning has also been widely used in multi-view clustering work, such as \cite{li2021contrastive, lin2021completer, roy2021self, lin2022contrastive}. \cite{lin2021completer} is base on information theory, which maximizes the mutual information between different views through contrastive learning, and solves the problem of view missing through a bidirectional prediction network.
A end to end online image clustering method was proposed in \cite{li2021contrastive}, where contrastive learning was used to explore the consistency information between clustering space and instance space.
A multi-view representation learning method \cite{hassani2020contrastive} is proposed to apply contrastive learning for solving graph classification problems.
\cite{xu2022multi} optimizes the objectives of different feature spaces separately through contrastive learning, and solves the conflict between view reconstruction loss and consistency loss.

\begin{table}[t]
\centering  % 显示位置为中间
\caption{Main symbols used in this paper.}
\label{tb0}
\begin{tabular}{l|l}
\toprule
Symbol             & Meaning                                                    \\ \midrule
m                  & The number of views.                                       \\ \midrule
N                  & The number of samples.                                     \\ \midrule
$\mathbf{X}^{(i)}$ & The original feature representation in $i$ view.           \\ \midrule
$\mathbf{Z}^{(i)}$ & The view-specific feature representation in $i$ view.      \\ \midrule
$\mathbf{S}^{(i)}$ & The semantic label of the $i$-th view.                     \\ \midrule
$\mathbf{Q}^{(i)}$ & The clustering allocation distribution of the $i$-th view. \\ \midrule
$D_{Z}$            & The dimensionality of  $\mathbf{Z}^{(i)}$                  \\ \midrule
$D_{i}$            & The dimensionality of  $\mathbf{X}^{(i)}$                  \\ \bottomrule
\end{tabular}
\end{table}
\section{Method}
In this section, we introduce the proposed method, termed Multi-level Consistency Collaborative Multi-view Clustering (MCoCo).
A multi-view dataset $\{\mathbf{X}^{(i)}\} _{i=1}^{m}$ with $m$ view, the $i$-th view is denoted by $\mathbf{X}^{(i)} \in \mathbf{R}^{N \times D_{i}}$, where $N$ denotes the number of samples and $D_{i}$ represents the dimension of the view. In order to be more clear and concise, we have listed the main symbols used in this article in Table \ref{tb0}.
MVC aims to partition the examples into $k$ clusters.

\subsection{Loss Function}

The loss function of MCoCo can be formulated as follows:
\begin{equation}
\label{ep1}
    \mathcal{L} = \mathcal{L}_{Re} + \mathcal{L}_{Co},
\end{equation}
where $\mathcal{L}_{Re}$ is the loss of within-view reconstruction and $\mathcal{L}_{Co}$ denotes multi-level collaborative learning loss.

\subsection{Within-view reconstruction}

Generally, the dimensions and input forms of different views of multi-view data are quite different, and at the same time, there may be some redundant information in the original data.
To learn a reliable representation for each view, we map the data of different views into a low-dimensional feature space by inputting the data $\mathbf{X}^{(i)}$ into the respective encoder $\bm{E}_{i}(\cdot;\theta _{i})$ with parameter $\theta _{i}$:
\begin{equation}
\label{ep2}
    \mathbf{Z}_{j}^{(i)}  =  \bm{E}_{i}(\mathbf{X}_{j}^{(i)};\theta _{i} ),
\end{equation}
where $\mathbf{X}_{j}^{(i)}$ is the $j$-th sample of $\mathbf{X}^{(i)}$ and $\mathbf{Z}_{j}^{(i)} \in \mathbf{R}^{D_{Z}}$ denotes the representation in the ${D_{Z}}$-dimensional feature space. Then we input this low-dimensional feature into the decoder $\bm{D}_{i}(\cdot;\phi _{i})$ with parameter $\phi _{i}$ for reconstruction:
\begin{equation}
\label{ep3}
    \mathbf{\hat{X}}_{j}^{(i)}  =  \bm{D}_{i}(\mathbf{Z}_{j}^{(i)};\phi  _{i} ),
\end{equation}
where $\mathbf{\hat{X}}_{j}^{(i)}$ is the reconstructed sample.
By minimizing the flowing reconstruction loss $\mathcal{L}_{Re}$, we can transform the input $\mathbf{X}^{(i)}$ into the representation $\mathbf{Z}^{(i)}$:
\begin{equation}
\label{ep4}
    \mathcal{L}_{Re} = \sum_{i=1}^{m} \sum_{j=1}^{N} ||\mathbf{X}_{j}^{(i)}-\bm{D}_{i}(\bm{E}_{i}(\mathbf{X}_{j}^{(i)};\theta _{i});\phi  _{i}) ||_{2}^{2}. 
\end{equation}

\subsection{Multi-level Collaborative Learning}

Based on the within-view reconstruction, we can obtain the low-dimensional representations $\{\mathbf{Z}^{(i)}\}_{i=1}^m$ of different views.
In order to use the multi-level consistency information of multi-view data to guide clustering, MCoCo achieves respective consistency goals in semantic and feature space, and makes multi-level spaces collaborate with each other.

Overall, the loss function $\mathcal{L}_{Co}$ of this section consists of two parts:
\begin{equation}
\label{ep5}
    \mathcal{L}_{Co} = \lambda _{1} \mathcal{L}_{Se} + \lambda _{2} \mathcal{L}_{Ml},
\end{equation}
where $\mathcal{L}_{Se}$ is the loss of semantic consistency, $\mathcal{L}_{Ml}$ indicates the multi-level consistency loss.
Regarding $\lambda _{1}$ and $\lambda _{2}$, they are two trade-off parameters.

\begin{algorithm}[t]
  \caption{Optimization algorithm of MCoCo}
  \label{alg1}
  % \LinesNumbered
  \begin{algorithmic}[1]
    \Require
      Multi-view dataset $\{\mathbf{X}^{(i)}\} _{i=1}^{m}$; Parameter $\tau$; Number of categories k.
    \Ensure
      Cluster assignment $\mathbf{Y}$.
    \State Initialize $\{\theta _{i}, \phi _{i}\}_{i=1}^m$ by minimizing Eq.\ref{ep4};
    \State Initialize views' cluster centroids $\{\mu^{(i)}\}_{i=1}^{m}$ by $k$-means.
    \While{not converged}
        \State  Obtain the view-specific representation $\{\mathbf{Z}^{(i)}\}_{i=1}^m$ through Eq.\ref{ep2};
        \State  Obtain semantic labels $\{\mathbf{S}^{(i)}\}_{i=1}^m$;
        \State  Obtain cluster assignments $\{\mathbf{Q}^{(i)}\}_{i=1}^m$ and it's target distribution $\{\mathbf{P}^{(i)}\}_{i=1}^m$ through Eq.\ref{ep8} and Eq.\ref{ep9};

        \State  Obtain the target distribution $\{\mathbf{S}^{'(i)}\}_{i=1}^m$ of semantic space through Eq.\ref{ep10};
        \State  Update $\{\theta _{i}, \phi _{i}\}_{i=1}^m$, $\varphi$ and $\{\mu^{(i)}\}_{i=1}^{m}$ with Eq.\ref{ep1};
    \EndWhile
  \end{algorithmic}
\end{algorithm}
\noindent {\bf Contrastive learning of semantic consistency}

Because the data of different views describe the same object, different views should have similar semantic labels.
Then the aligned semantic labels can be used as a self-supervised signal to modify the cluster assignments in feature space, which makes the obtained cluster assignments more discriminating.
In this section, we explain how to obtain consistent semantic labels.
Specifically, We obtain the semantic labels of each view $\{\mathbf{S}^{(i)} \in \mathbf{R}^{N \times k} \}_{i=1}^m$ by inputing $\{\mathbf{Z}^{(i)}\}_{i=1}^m$ into a shared semantic generator $\bm{G}_c(\cdot;\varphi)$ with the parameter $\varphi$,
which is constructed by fully connected neural networks.
The $\mathbf{S}_{ij}^{(m)}$ represents the probability that the $i$-th sample in view $m$ belongs to the $j$-th class.
To effectively excavate the variable semantic consensus information in semantic space and make semantic labels of different views tend to be consistent, the contrastive learning of semantic consistency is introduced here.
For $\mathbf{S}_{\cdot j}^{(m)}$, there are $(mk - 1)$ column vector pairs $\{\mathbf{S}_{\cdot j}^{(m)}, \mathbf{S}_{\cdot c}^{(w)}\}_{c=1,...,k}^{w=1,...,m}$, of which $\{\mathbf{S}_{\cdot j}^{(m)}, \mathbf{S}_{\cdot j}^{(w)}\}_{w \neq m}$ can form $(m-1)$ positive pairs and the remaining $m(k-1)$ pairs form negative pairs.
The cosine similarity is utilized to measure the similarity between two semantic column vectors:
\begin{equation}
\label{ep6}
    d(\mathbf{S}_{\cdot j}^{(i)}, \mathbf{S}_{\cdot c}^{(j)}) = \frac{\mathbf{S}_{\cdot i}^{(i)} \cdot \mathbf{S}_{\cdot c}^{(j)}}{||\mathbf{S}_{\cdot i}^{(i)}||||\mathbf{S}_{\cdot c}^{(j)}||}.  
\end{equation}

We define the semantic consistency loss $l{(i, j)}$ between $\mathbf{S}^{(i)}$ and $\mathbf{S}^{(j)}$ as:
\begin{equation}
     -\frac{1}{k}\sum_{c=1}^{k}  \log_{}{\frac{e^{d(\mathbf{S}_{\cdot c}^{(i)},\mathbf{S}_{\cdot c}^{(j)})/\tau } }{ (\sum_{w=1}^{k}(e^{d(\mathbf{S}_{\cdot c}^{(i)}, \mathbf{S}_{\cdot w}^{(i)})/\tau }+e^{d(\mathbf{S}_{\cdot c}^{(i)}, \mathbf{S}_{\cdot w}^{(j)})/\tau })-e^{\frac{1}{\tau} }} \nonumber},
\end{equation}
where $\tau$ is the temperature parameter.
The complete contrastive learning loss of semantic consistency can be formulated as follows:
\begin{equation}
\begin{split}
\label{ep7}
    \mathcal{L}_{Se} &= \frac{1}{2}\sum_{i=1}^{m}\sum_{j=1, j \neq i}^{m} l(i, j)\\
    &+ \sum_{i=1}^{m}\sum_{c=1}^{k} (\frac{1}{N}\sum_{j=1}^{N}\mathbf{S}_{jc}^{(i)} \log_{}{\frac{1}{N}\sum_{j=1}^{N}\mathbf{S}_{jc}^{(i)}}  ).
\end{split}
\end{equation}

The second part of Eq.\ref{ep7} is a regularization term, which can avoid grouping all samples into the same cluster.

\noindent {\bf Multi-level consistency collaboration}

In order to obtain the clustering assignments of each view in feature space, we initialize the learnable parameters $\{\mu_{j}^{(i)} \in \mathbf{R}^{D_{Z}}\}_{j=1}^k$ by $k$-means\cite{xie2016unsupervised}, where $\mu_{j}^{(i)}$ represents the $j$-th cluster centroid of the $i$-th view.
According to \cite{xie2016unsupervised,xu2022self,li2019deep}, we use Student's $t$-distribution to generate soft cluster assignments, which can be described as:
\begin{equation}
\label{ep8}
\mathbf{Q}_{ij}^{(m)} = \frac{(1 + ||\mathbf{Z}_{i}^{(m)} - \mu_j  ^{(m)}||^2)^{-1}}{ {\textstyle \sum_{j}^{}(1 + ||\mathbf{Z}_{i}^{(m)} - \mu_j  ^{(m)}||^2)^{-1}} },
\end{equation}
where $\mathbf{Q}_{ij}^{(m)}$ is treated as the pseudo label that represents the probability of assigning the $i$-th sample of the $m$-th view to the $j$-th category.
The higher probability of pseudo label means that the  high probability components of pseudo label has higher confidence. In order to increase the discrimination ability of pseudo label with higher confidence, we enhance $\mathbf{Q}_{ij}^{(m)}$ to an auxiliary target distribution $\mathbf{P}^{(m)}$ with the operation of square and normalization:
\begin{equation}
\label{ep9}
\mathbf{P}_{ij}^{(m)} = \frac{(\mathbf{Q}_{ij}^{(m)})^{2}/ {\textstyle \sum_{i}^{}}\mathbf{Q}_{ij}^{(m)}  }{ {\textstyle \sum_{j}^{}((\mathbf{Q}_{ij}^{(m)})^{2}/ {\textstyle \sum_{i}^{}}\mathbf{Q}_{ij}^{(m)} )} }.
\end{equation}

\begin{table*}[t]
\centering  % 显示位置为中间
\caption{The information of the datasets in our experiments.}
\label{tb1}
\small
\resizebox{0.75\textwidth}{!}{
% \small
\begin{tabular}{lllll}
\toprule
Dataset       & \#Sample & \#Cluster & \#View & \#Dimensionality of features  \\ \midrule
MNIST-USPS    & 5000     & 10        & 2      & \{784, 784\}          \\
Multi-COIL-20 & 1440     & 20        & 3      & \{1024, 1024, 1024\} \\
BDGP          & 2500     & 5         & 2      & \{1750, 79\}                  \\
Multi-MNIST   & 70000    & 10        & 2      & \{1024, 1024\}          \\
Multi-Fashion & 10000    & 10        & 3      & \{784, 784, 784\} \\
Noisy-MNIST   & 50000    & 10        & 2      & \{1024, 1024\}          \\
Caltech-2V    & 1400     & 7         & 2      & \{40, 254\}                   \\
Caltech-3V    & 1400     & 7         & 3      & \{40, 254, 928\}              \\
Caltech-4V    & 1400     & 7         & 4      & \{40, 254, 928, 512\}         \\
Caltech-5V    & 1400     & 7         & 5      & \{40, 254, 928, 512, 1984\}   \\ \bottomrule
\end{tabular}}
\end{table*}
% \begin{algorithm}[t]
%   \caption{Optimization algorithm of MCoCo}
%   \label{alg1}
%   % \LinesNumbered
%   \begin{algorithmic}[1]
%     \Require
%       Multi-view dataset $\{\mathbf{X}^{(i)}\} _{i=1}^{m}$; Parameter $\tau$; Number of categories k.
%     \Ensure
%       Cluster assignment $\mathbf{Y}$.
%     \State Initialize $\{\theta _{i}, \phi _{i}\}_{i=1}^m$ by minimizing Eq.\ref{ep4};
%     \State Initialize views' cluster centroids $\{\mu^{(i)}\}_{i=1}^{m}$ by $k$-means.
%     \While{not converged}
%         \State  Obtain the view-specific representation $\{\mathbf{Z}^{(i)}\}_{i=1}^m$ through Eq.\ref{ep2};
%         \State  Obtain semantic labels $\{\mathbf{S}^{(i)}\}_{i=1}^m$;
%         \State  Obtain cluster assignments $\{\mathbf{Q}^{(i)}\}_{i=1}^m$ and it's target distribution $\{\mathbf{P}^{(i)}\}_{i=1}^m$ through Eq.\ref{ep8} and Eq.\ref{ep9};

%         \State  Obtain the target distribution $\{\mathbf{S}^{'(i)}\}_{i=1}^m$ of semantic space through Eq.\ref{ep10};
%         \State  Update $\{\theta _{i}, \phi _{i}\}_{i=1}^m$, $\varphi$ and $\{\mu^{(i)}\}_{i=1}^{m}$ with Eq.\ref{ep1};
%     \EndWhile
%   \end{algorithmic}
% \end{algorithm}
In order to make multi-views collaborate with each other to get consistent cluster assignments and, at the same time, make use of the aligned semantic labels to weakly supervise the cluster assignments in feature space, we propose a brand-new multi-level consistency collaboration strategy, which can achieve multi-level collaboration.
Specifically, we enhance the semantic labels of each view according to Eq.\ref{ep9} to get the target distribution of semantic space:
\begin{equation}
\label{ep10}
\mathbf{S}^{'(m)}_{ij} = \frac{(\mathbf{S}_{ij}^{(m)})^{2}/ {\textstyle \sum_{i}^{}}\mathbf{S}_{ij}^{(m)}  }{ {\textstyle \sum_{j}^{}((\mathbf{S}_{ij}^{(m)})^{2}/ {\textstyle \sum_{i}^{}}\mathbf{S}_{ij}^{(m)} )} }.
\end{equation}

\noindent As thus, the multi-level consistency loss $\mathcal{L}_{Ml}$ is defined as:
\begin{equation}
\begin{split}
\label{ep11}
\mathcal{L}_{Ml} &=  {\sum_{k=1}^{m}}(\sum_{c=1}^{m} (D_{kl}(\mathbf{P}^{(c)}||\mathbf{Q}^{(k)})) + D_{kl}(\mathbf{S}^{'(k)}||\mathbf{Q}^{(k)}))\\
&=  {\sum_{k=1}^{m}}{\sum_{i=1}^{N}} { \sum_{j=1}^{k}} (\sum_{c=1}^{m}(\mathbf{P}^{(c)}_{ij}\log_{}{\frac{\mathbf{P}^{(c)}_{ij}}{\mathbf{Q}_{ij}^{(k)}}}) +\mathbf{S}_{ij}^{'(k)}\log_{}{\frac{\mathbf{S}^{'(k)}_{ij}}{\mathbf{Q}_{ij}^{(k)}}}), 
\end{split}
\end{equation}
where $D_{kl}$ indicates the Kullback-Leibler divergence.
By optimizing Eq.\ref{ep11}, different views jointly learned with each other in the feature space, and the samples with similar semantic information attract each other.
In this way, MCoCo can mine multi-level consistency information to learn discriminating clustering assignments.

MCoCo's multi-level consistency collaboration strategy enables it to obtain the consistent cluster assignments of multiple views.
In order to avoid the interference of a few false predictions and realize clear cluster assignments with high confidence, the final cluster assignment is calculated as follows:
\begin{equation}
\label{ep12}
\mathbf{Y}_{i} = \mathrm{arg}\mathop{\mathrm{max}}\limits_{j}({\frac{1}{m}\sum_{k=1}^{m}\mathbf{Q}_{ij}^{(k)} }).
\end{equation}

For clarification, the optimization procedure of MCoCo is summarized in Algorithm \ref{alg1}.

\section{Experiments}
To verify the effectiveness of our method, extensive experiments are conducted in this section.
Furthermore, detailed discussions of our method are provided as well.

\subsection{Experiments Setup}
\noindent {\bf Datasets.}
The benchmark datasets we used in our experiments are shown in Table \ref{tb1}:

{\bf 1) MNIST-USPS} \cite{peng2019comic}: It is a two-view dataset that contains 5000 handwritten digital image samples from numbers 0 to 9.

{\bf 2) Multi-COIL-20} \cite{wan2021multi}: It is a three-view dataset containing 1440 pictures of 20 categories, and different views represent different poses of the same object.

{\bf 3) BDGP} \cite{cai2012joint}: It is a two-view dataset containing 2500 images of drosophila embryos belonging to 5 categories, each with visual and textual features.

{\bf 4) Multi-MNIST} \cite{xu2021multi}: It contains 70000 handwritten digital images belonging to 10 classes, which have two views, and different views imply the same digit written by different people.

{\bf 5) Multi-Fashion} \cite{xiao2017fashion}: It has 10000 images collected from 10 categories about fashion products and has three views.
We use the same data set as \cite{xu2022self}, which randomly selects a sample with the same label from this set to construct the second and third view.

{\bf 6) Noisy-MNIST} \cite{lin2021completer}: It uses the original 70000 MNIST images as the first view and randomly selects within-class images with white Gaussian noise as the second view.
We followed the setting of article \cite{xu2019learning} and used a subset of Noisy-MNIST containing 50000 samples.

{\bf 7) Caltech-$n$V} \cite{fei2004learning}: It is an RGB image dataset with multiple views, which contains 1400 images belonging to 7 categories and have five views.
Four sub-datasets, namely Caltech-2V, Caltech-3V, Caltech-4V, and Caltech-5V, with different numbers of views, are built for evaluating the robustness of the comparison methods in terms of the number of views. Specifically, Caltech-2V uses WM and CENTRISTT; Caltech-3V uses WM, CENTRIST, and LBP; Caltech-4V uses WM, CENTRIST, LBP, and GIST; Caltech-5V uses WM, CENTRIST, LBP, GIST, and HOG.

% 小数据表
\begin{table*}[t]
\centering  % 显示位置为中间
\caption{Clustering results on small-scale datasets. The best and second best results have been marked in bold and underlined respectively.}  %
\label{tb2}
\resizebox{0.98\textwidth}{!}{
\begin{tabular}{r|cccc|cccc|cccc}
\toprule
Datasets                              & \multicolumn{4}{c|}{MNIST-USPS}                                                                                                       & \multicolumn{4}{c|}{Multi-COIL-20}                                                                                                    & \multicolumn{4}{c}{BDGP}                                                                                                              \\ \midrule
Metrics                               & ACC                             & NMI                             & Fscore                          & RI                              & ACC                             & NMI                             & Fscore                          & RI                              & ACC                             & NMI                             & Fscore                          & RI                              \\ \midrule
LMSC(2017)                            & 0.373                           & 0.433                           & 0.404                                & 0.771                                & 0.683                                & 0.766                                & 0.613                                & 0.954                                & 0.524                               & 0.432                                & 0.556                                & 0.604                                \\
AE$^2$-Nets(2019)                     & 0.626                                & 0.623                                & 0.567                                & 0.903                                & 0.740                                & 0.862                                & 0.771                                & 0.964                                & 0.552                                & 0.406                                & 0.501                                & 0.661                                \\
DEMVC(2021)                           & 0.901                                & 0.930                                & 0.897                                & 0.979                                & \underline{0.825}                                & \underline{0.935}                                & \underline{0.871}                                & 0.980                                & 0.609                                & 0.529                                & 0.557                                & 0.818                                \\
DUA-Nets(2021)                        & 0.751                                & 0.689                                & 0.930                                & 0.660                                & 0.602                                & 0.711                                & 0.597                                & 0.946                                & 0.603                                & 0.406                                & 0.539                                & 0.762                                \\
DCP(2022)                             & 0.891                                & 0.941                                & \underline{0.928}                                & 0.976                                & 0.690                                & 0.887                                & 0.621                                & 0.958                                & 0.438                                & 0.385                                & 0.534                                & 0.542                                \\
SDMVC(2022)                           & \underline{0.937}                                & \underline{0.943}                                & 0.913                                & \underline{0.983}                                & 0.809                           & 0.905                           & 0.823                           & \underline{0.981}                           & \underline{0.965}                           & \underline{0.909}                           & \underline{0.934}                           & \underline{0.961}                           \\
CMRL(2023)                            & 0.916                                & 0.856                                & 0.860                                & 0.971                                & 0.792                                & 0.878                                & 0.797                                & 0.977                                & 0.789                                & 0.672                                & 0.826                                & 0.717                                \\
\textbf{MCoCo(ours)} & \textbf{0.995}          & \textbf{0.986}          & \textbf{0.998}          & \textbf{0.990}              & \textbf{0.999}          & \textbf{0.999}          & \textbf{0.999}          & \textbf{0.999}              & \textbf{0.987}          & \textbf{0.959}          & \textbf{0.989}          & \textbf{0.972}             \\ \bottomrule
\end{tabular}}
\end{table*}
\begin{figure*}[h]
\centering
\subfloat[Epoch 0]{
\includegraphics[width=0.18\linewidth]{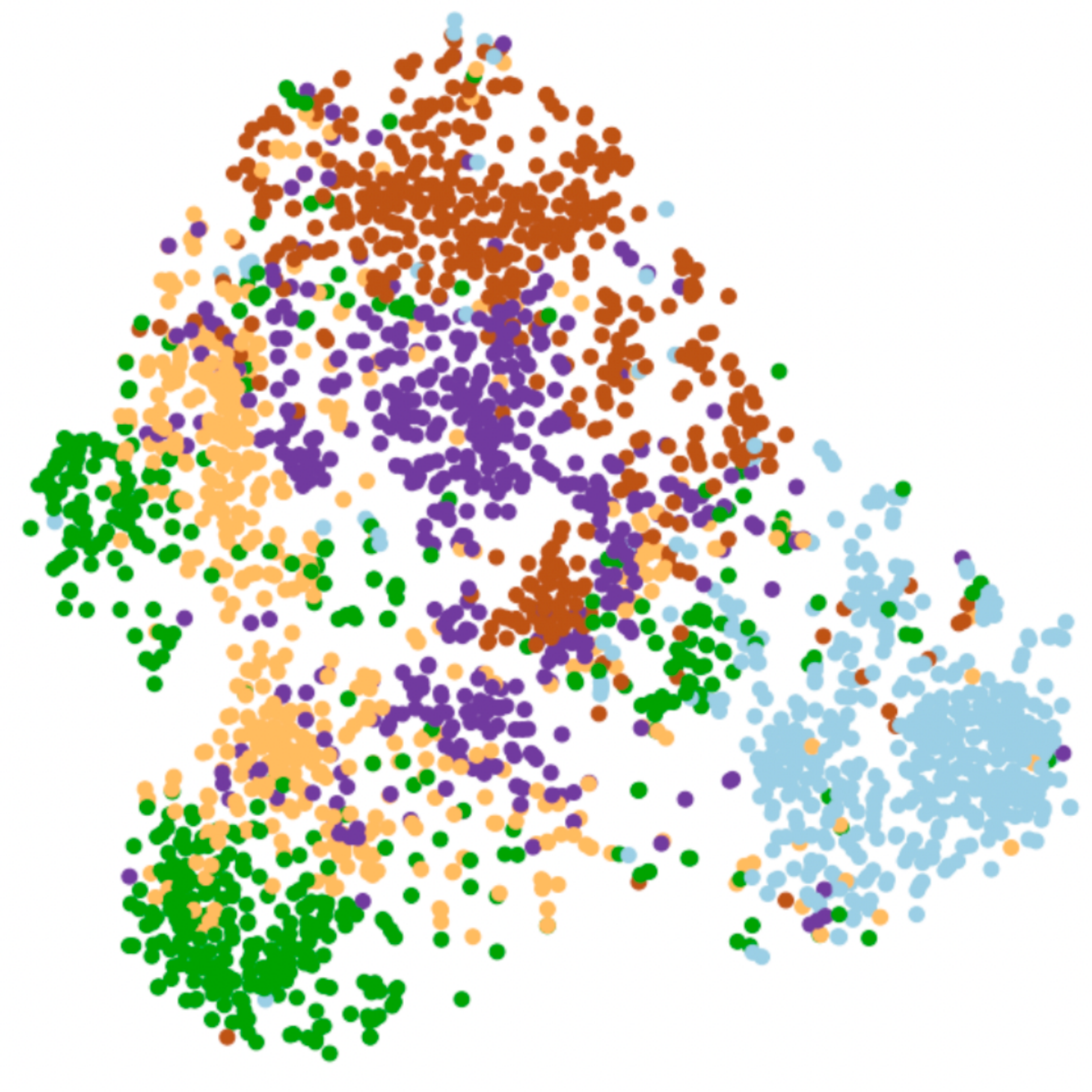}
\captionsetup{font=small}

}
\hfill
\subfloat[Epoch 10]{
\includegraphics[width=0.18\linewidth]{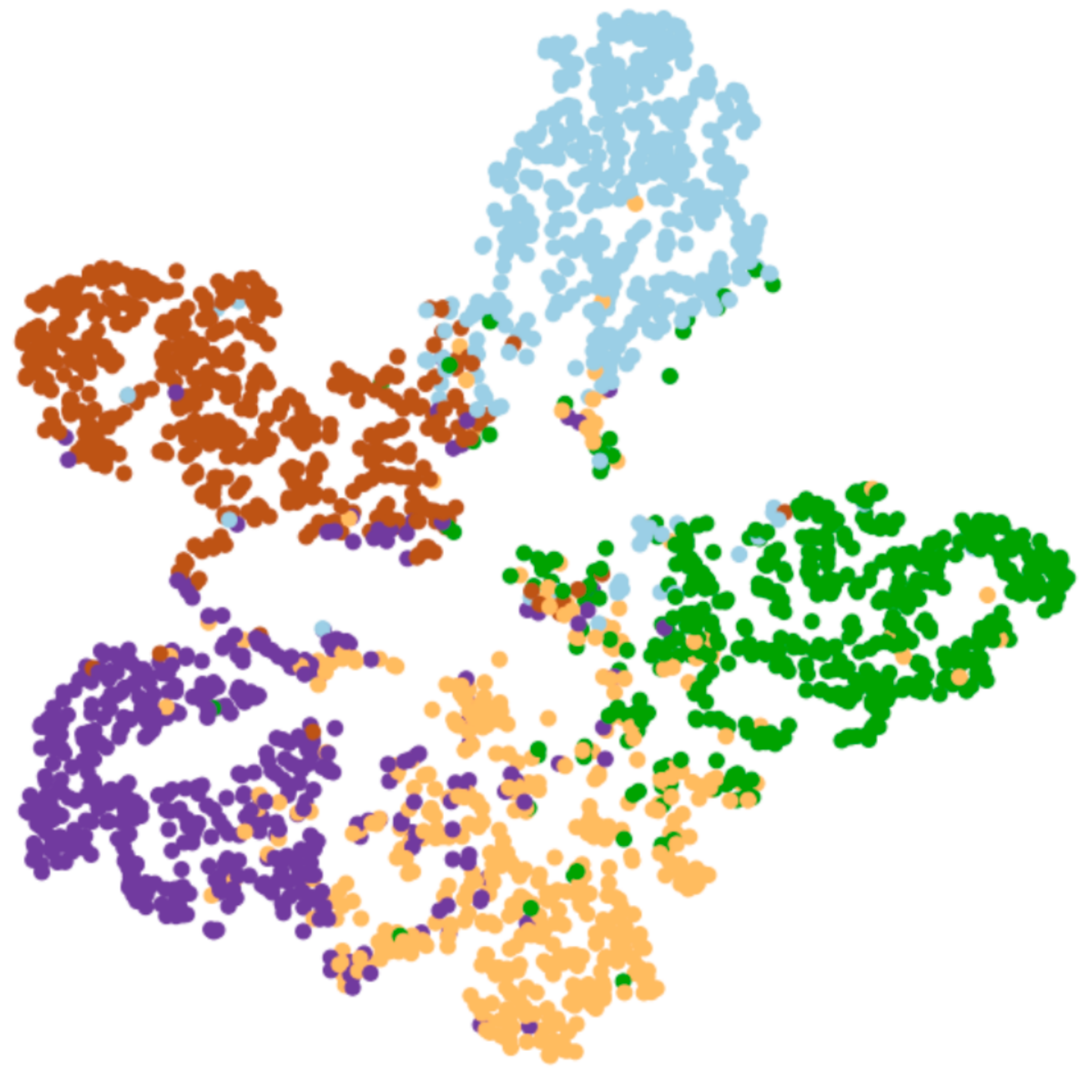}
\captionsetup{font=small}
}
\hfill
\subfloat[Epoch 20]{
\includegraphics[width=0.18\linewidth]{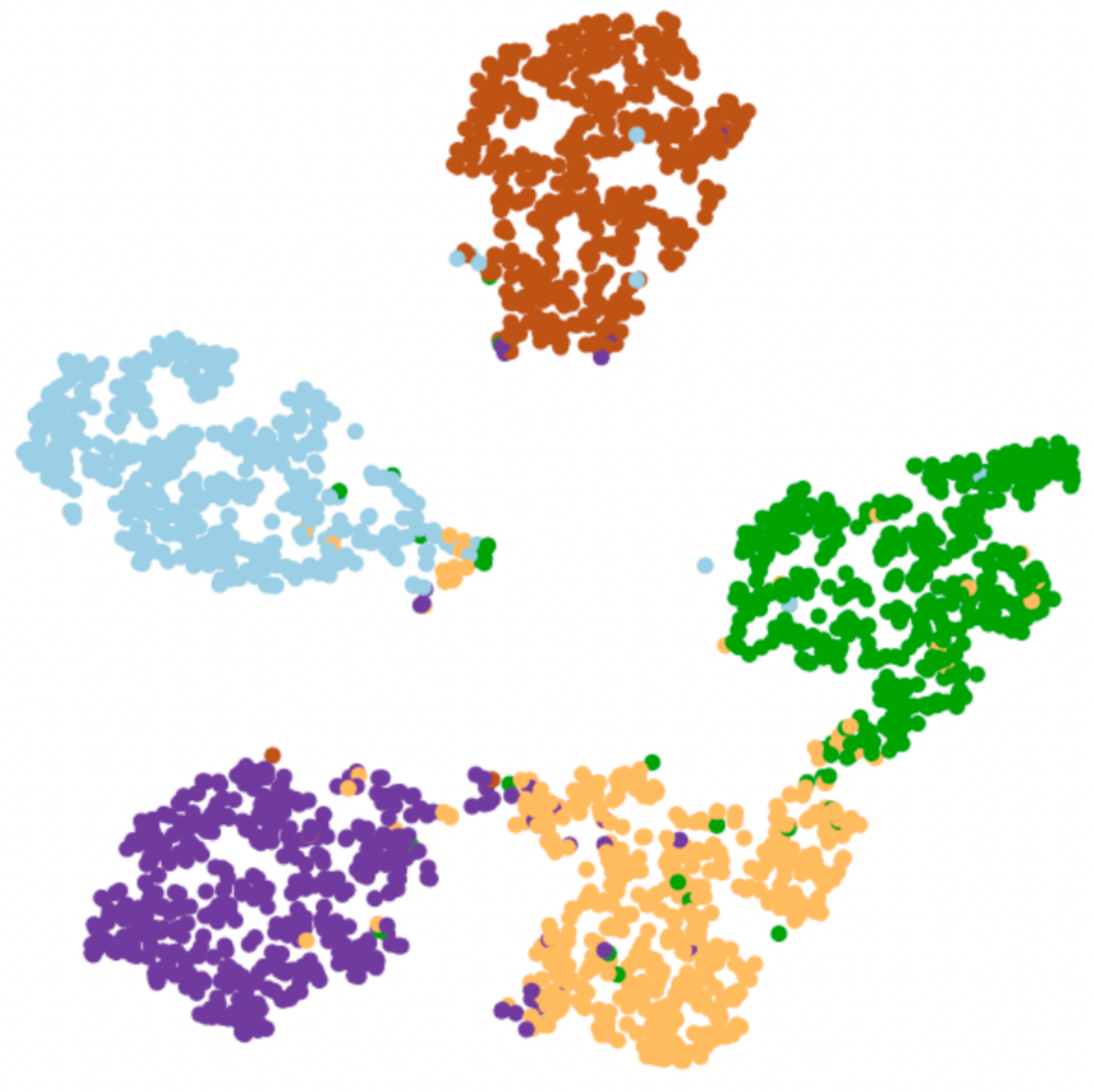}
\captionsetup{font=small}
}
\hfill
\subfloat[Epoch 30]{
\includegraphics[width=0.18\linewidth]{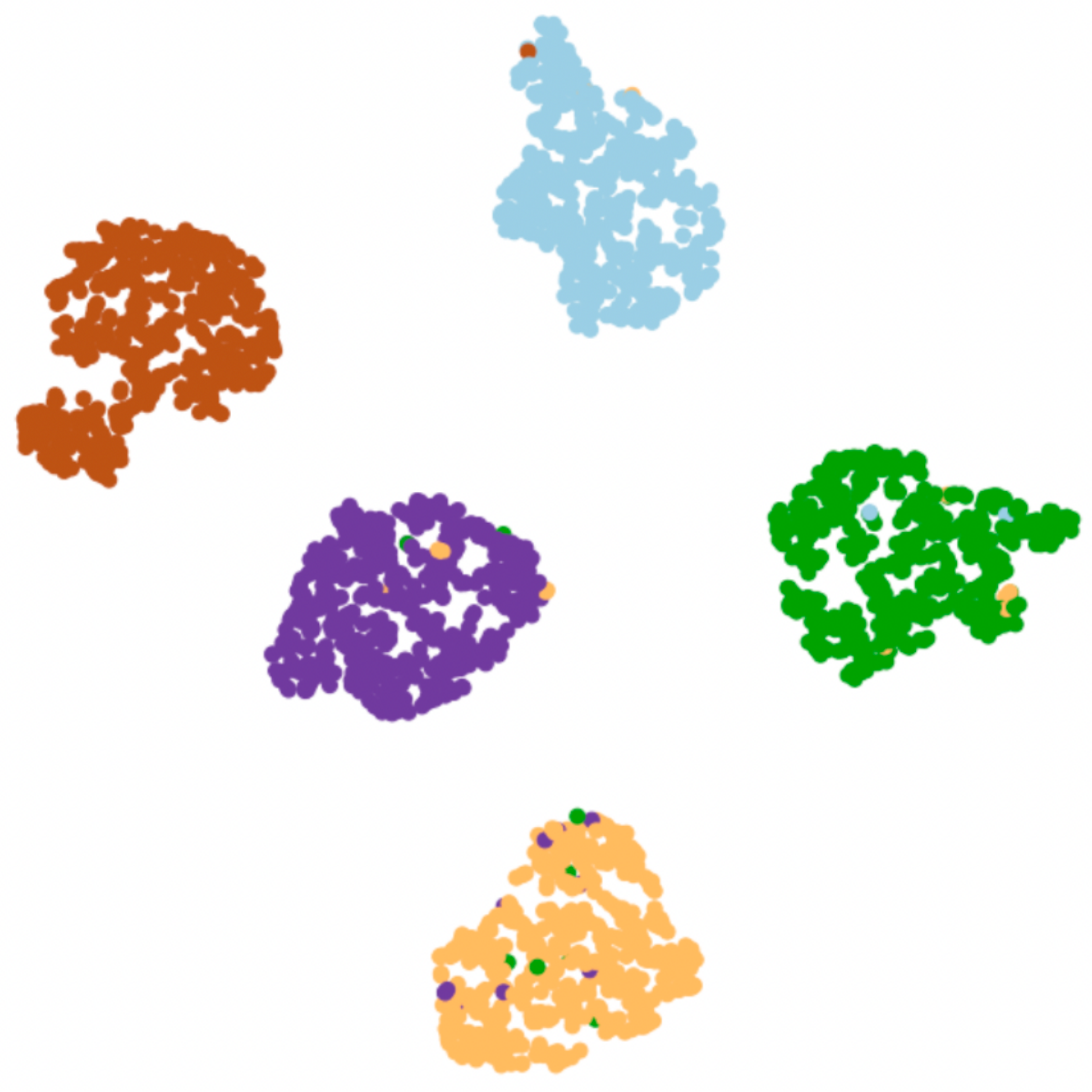}
\captionsetup{font=small}
}
\quad
\subfloat[Epoch 30(w/o Ml)]{
\includegraphics[width=0.18\linewidth]{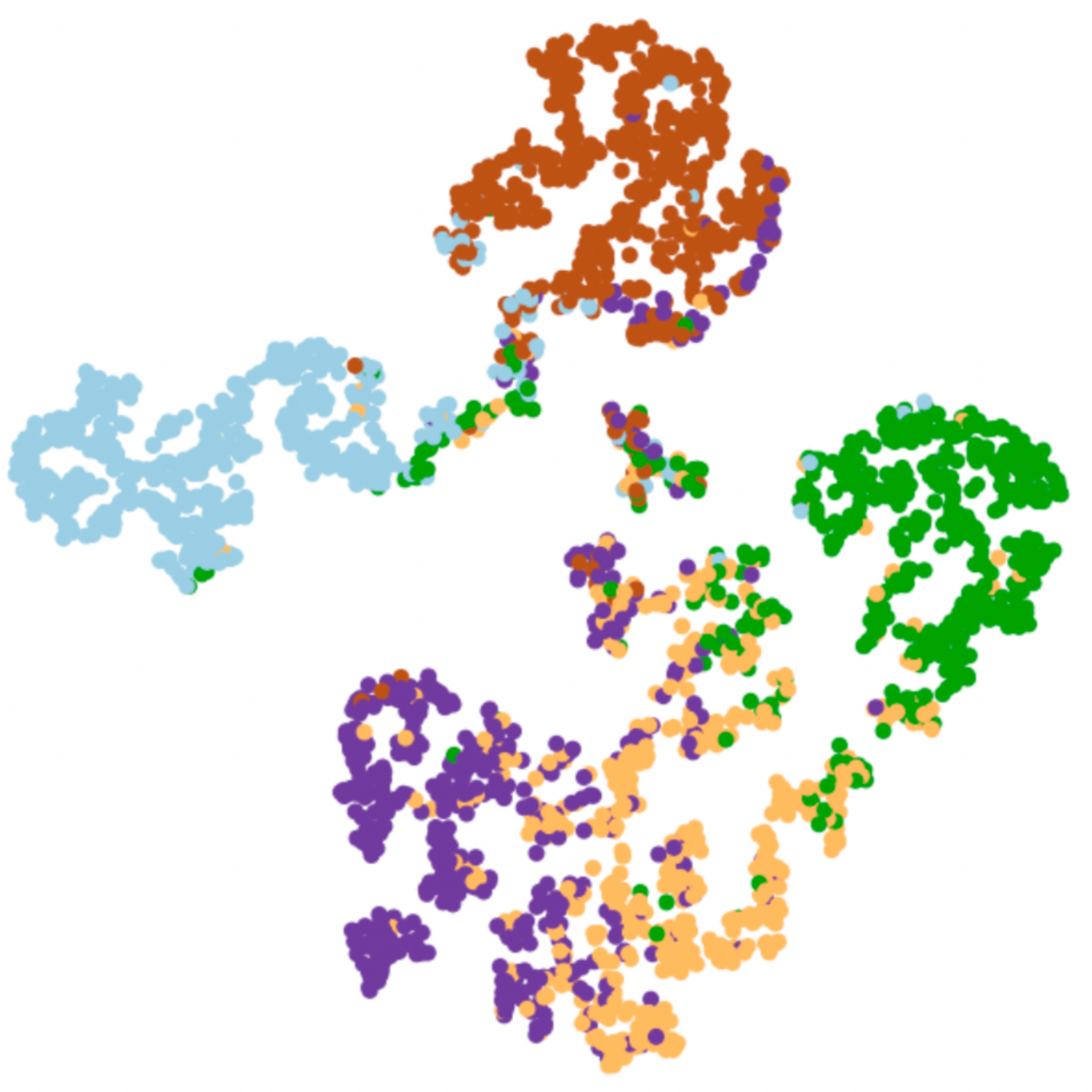}
\captionsetup{font=small}
}
\quad
\caption{The t-SNE visualization results of the BDGP dataset (view one) in Epoch 0, 10, 20, and 30. (e) showing the representation of view one learned by MCoCo at epoch 30 without multi-level collaboration.}
\label{fg3}
\end{figure*}
\noindent{\bf Evaluation metrics.}
Four metrics are utilized to evaluate the clustering quality, i.e., Accuracy (ACC), Normalized Mutual Information (NMI),  Rand Index (RI), and Fscore.
To eliminate the randomness and make the experimental results more reliable, we take ten trials for all experiments.

\noindent{\bf Comparison methods.}
The following state-of-the-art algorithms are used for comparison:

{\bf 1) LMSC} \cite{zhang2017latent}: It learns the latent unified representation by mapping different views' view-specific feature into a common space and employing the low-rank subspace constraint.

{\bf 2) AE$^2$-Nets} \cite{zhang2019ae2}: Nested autoencoders are used to learn the compact unified representation by balancing the complementarity and consistency among multiple views.

{\bf 3) DEMVC} \cite{xu2019learning}: It proposed a non-fusion model of collaborative training among cluster assignments of multiple views.

{\bf 4) DUA-Nets} \cite{geng2021uncertainty}: It presents the dynamic uncertainty-aware networks for UMRL.
By estimating and leveraging the uncertainty of data, it achieves the noise-free multi-view feature representation.

{\bf 5) DCP} \cite{lin2022dual}: It learns the unified multi-view representation by maximizing the mutual information of different views via contrastive learning in the feature space.

{\bf 6) CMRL} \cite{zheng2023comprehensive}: It introduces the orthogonal mapping strategy and imposing the low-rank tensor constraint on the subspace representations.

{\bf 7) SDMVC} \cite{xu2022self}: It concatenates the features of multiple views as global feature and uses global discriminative information to supervise all views to learn more discriminative view-specific features.

% 大数据
\noindent{\bf Implementation details.} For all datasets, the ReLU \cite{glorot2011deep} activation function is used to implement autoencoders in MCoCo.
Adam optimizer \cite{kingma2014adam} is employed for optimization.
Our method is implemented by PyTorch \cite{paszke2019pytorch} on one NVIDIA Geforce GTX 2080ti GPU with 11GB memory.

\subsection{Experimental Result}
We discuss the clustering performance of MCoCo compared with other state-of-art algorithms on three different datasets: small-scale datasets, large-scale datasets, and datasets with a variable number of views. Generally speaking, the proposed method can achieve the best performance in all cases.

\begin{table*}[t]
\centering  % 显示位置为中间
\caption{Clustering results on large-scale datasets. Due to the high complexity of some methods, the unknown value is represented by "-".} 
\label{tb3}
\resizebox{0.98\textwidth}{!}{
\begin{tabular}{r|cccc|cccc|cccc}
\toprule
Datasets             & \multicolumn{4}{c|}{Multi-MNIST}                                                                          & \multicolumn{4}{c|}{Multi-Fashion}                                                                        & \multicolumn{4}{c}{Noisy-MNIST}                                                                          \\ \midrule
Metrics   & ACC & NMI & Fscore & RI & ACC & NMI & Fscore & RI & ACC & NMI & Fscore & RI \\ \midrule
LMSC(2017)       & -                        & -                        & -                        & -                            & 0.439                        & 0.417                        & 0.367                        & 0.828                            & -                        & -                        & -                        & -                            \\
AE$^2$-Nets(2019)           & 0.737                        &    0.645                     &   0.624                      & 0.924                            & 0.729                        & 0.763                        & 0.716                        & 0.935                            & 0.220                        & 0.098                        & 0.162                        & 0.813                           \\
DEMVC(2021)       & 0.996                        & \underline{0.997}                        & 0.996                        & 0.996                            & 0.720                        & 0.848                        & 0.778                        & 0.926                            & 0.589                        & 0.714                        & 0.633                        & 0.900                           \\
DUA-Nets(2021)          & 0.795                        & 0.742                        & 0.713                        & 0.943                            & 0.772                        & 0.761                        & 0.727                        & 0.945                            & 0.179                        & 0.066                        & 0.145                        & 0.808                         \\
DCP(2022)          & 0.793                        & 0.905                        & 0.747                        & 0.952                            & 0.757                        & 0.862                        & 0.822                        & 0.948                            & \underline{0.786}                        & \underline{0.883}                        & \underline{0.740}                        & \underline{0.921}                           \\
SDMVC(2022)          & \underline{0.998}                   & 0.996                   & \underline{0.998}                   & \underline{0.998}                       & \underline{0.860}                   & \underline{0.876}                   & \underline{0.845}                   & \underline{0.965}                       & 0.557                        & 0.527                        & 0.460                        & 0.887                           \\
CMRL(2023)           & -                       & -                       & -                       & -                           & 0.768                   & 0.803                   & 0.748                   & 0.943                       & -                       & -                       & -                       & -                          \\
\textbf{MCoCo(ours)} & \textbf{0.999}          & \textbf{0.998}          & \textbf{0.999}          & \textbf{0.999}              & \textbf{0.991}          & \textbf{0.977}          & \textbf{0.996}          & \textbf{0.982}              & \textbf{0.994}          & \textbf{0.981}          & \textbf{0.998}          & \textbf{0.988}             \\ \bottomrule
\end{tabular}}
\end{table*}
\begin{figure*}[h]
\centering
\subfloat[MNIST-USPS]{
\includegraphics[width=0.18\linewidth]{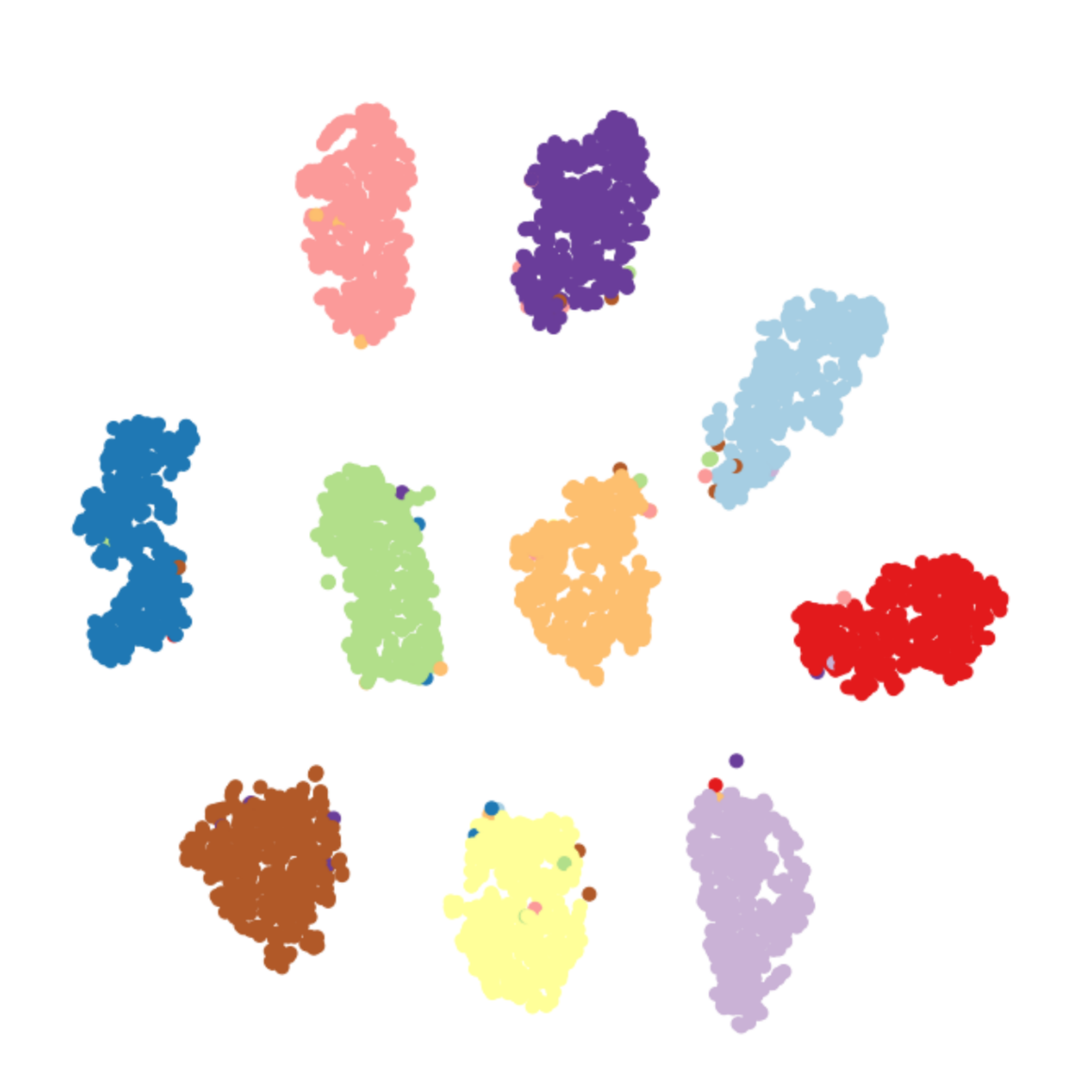}
\captionsetup{font=small}

}
\hfill
\subfloat[Multi-COIL-20]{
\includegraphics[width=0.18\linewidth]{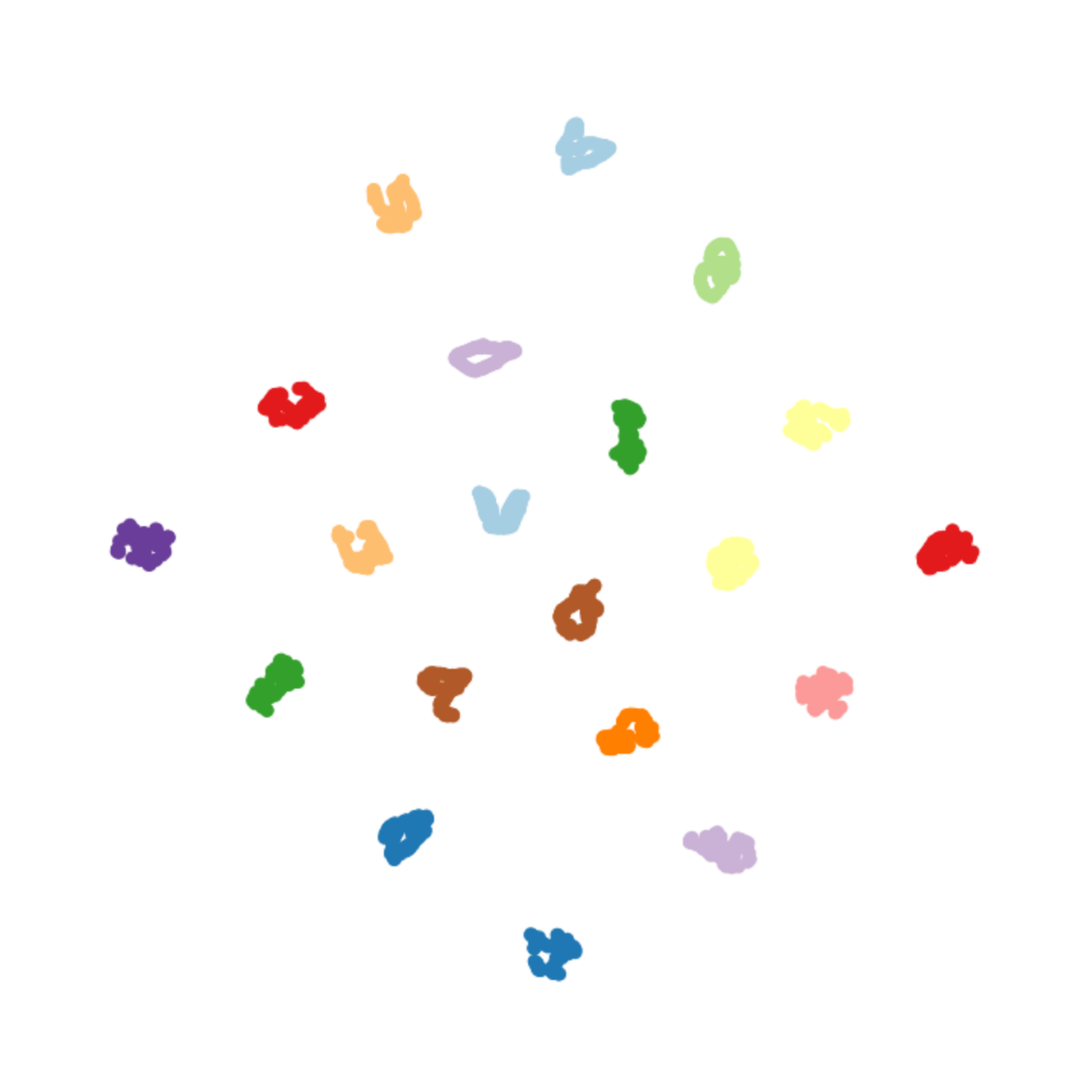}
\captionsetup{font=small}
}
\hfill
\subfloat[Multi-MNIST]{
\includegraphics[width=0.18\linewidth]{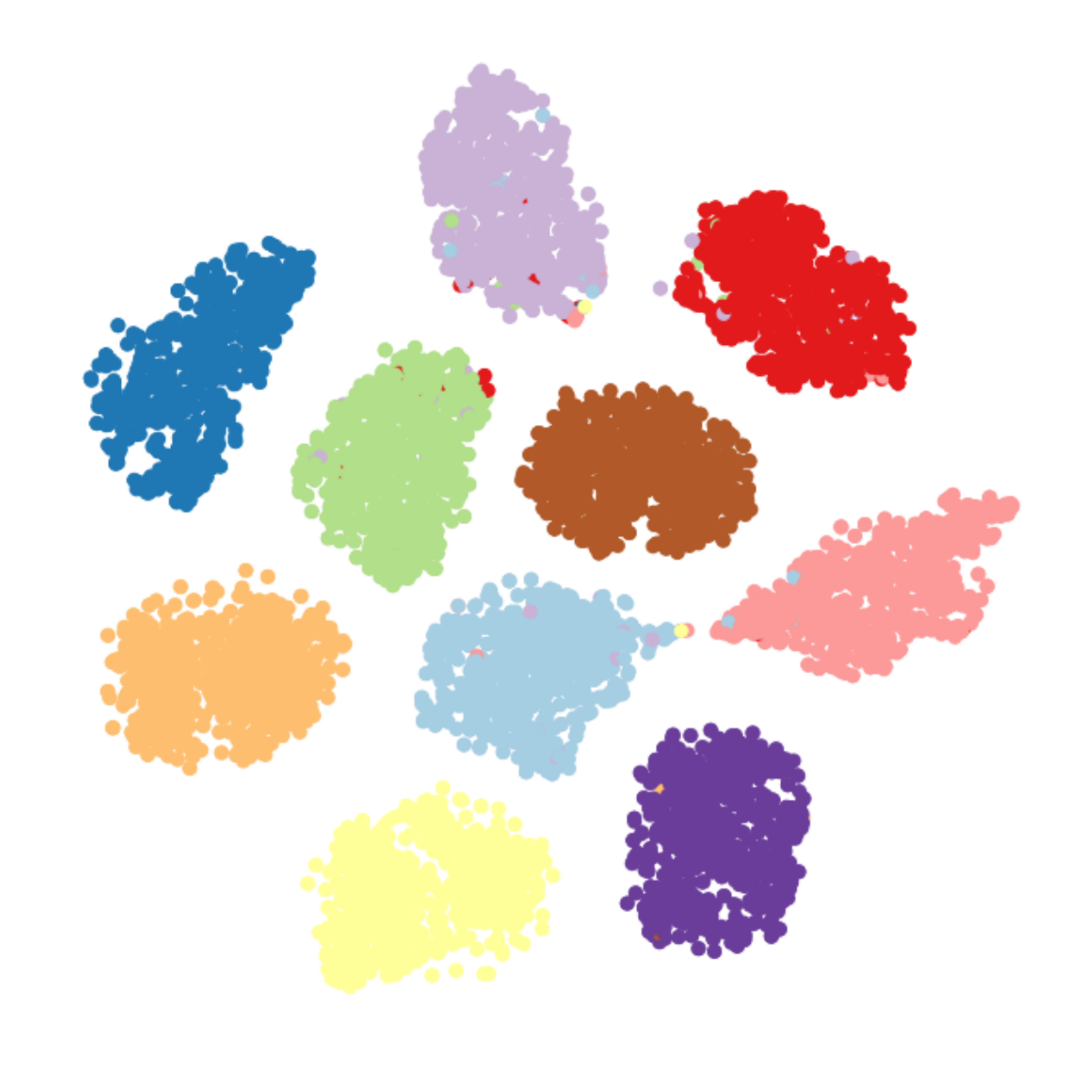}
\captionsetup{font=small}
}
\hfill
\subfloat[Multi-Fashion]{
\includegraphics[width=0.18\linewidth]{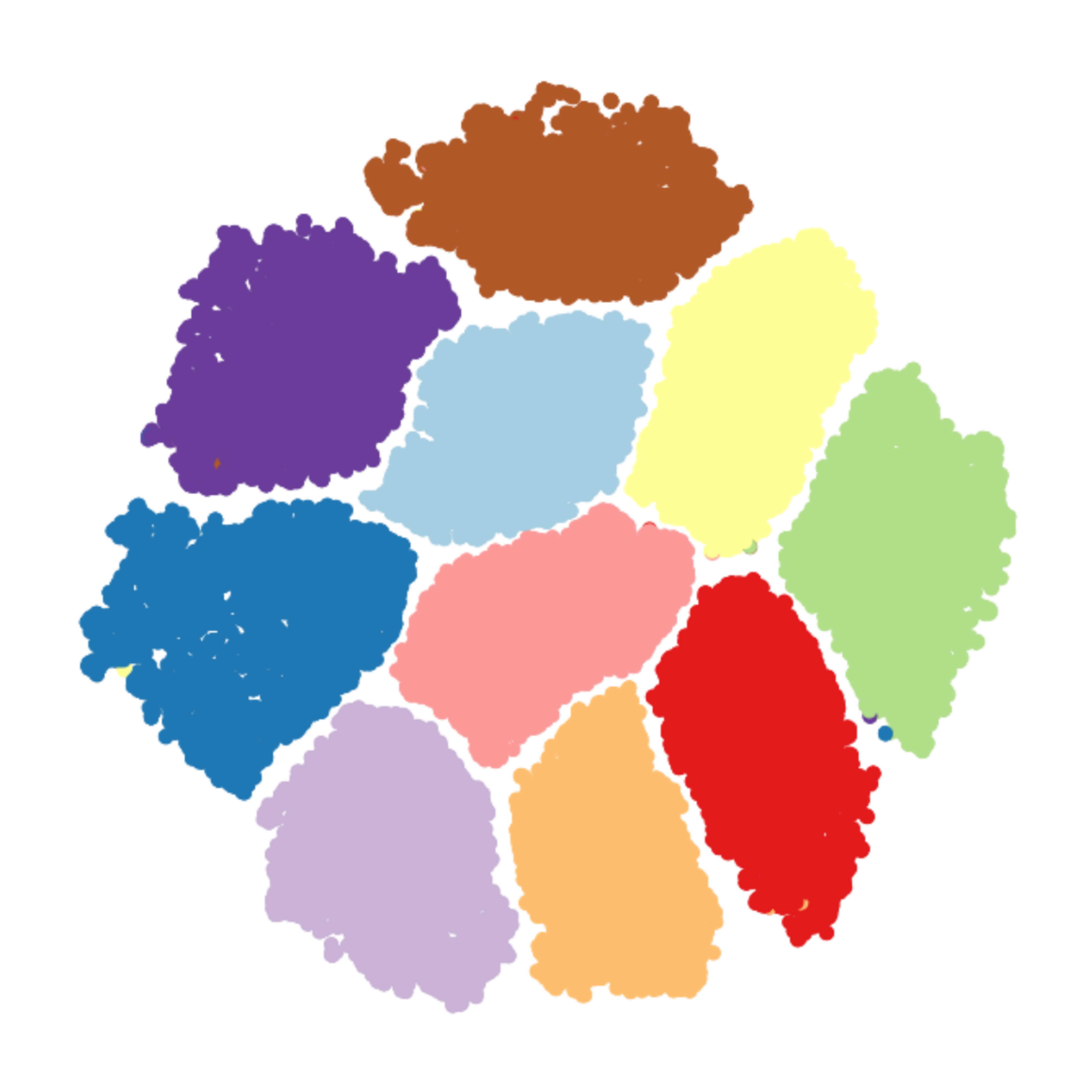}
\captionsetup{font=small}
}
\quad
\subfloat[Noisy-MNIST]{
\includegraphics[width=0.18\linewidth]{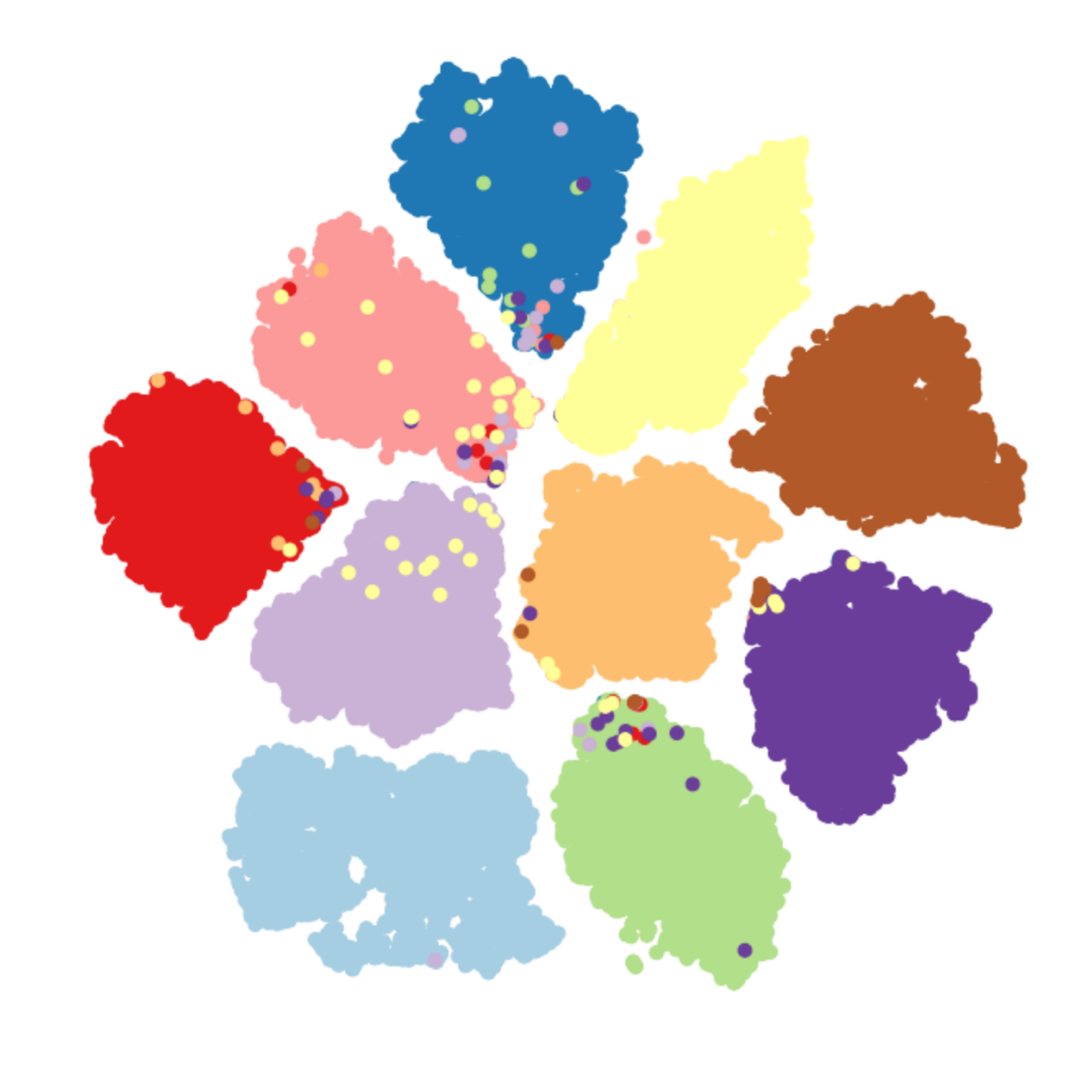}
\captionsetup{font=small}
}
\quad
\caption{The t-SNE visualization results of view one in MNIST-USPS, Multi-COIL-2O, Multi-MNIST, Multi-Fashion and Noisy-MNIST.}
\label{fg4}
\end{figure*}
The results on small-scale and large-scale datasets are reported in Tables \ref{tb2} and \ref{tb3}.
We can observe that MCoCo has achieved the best performance on all metrics, whether it is a large-scale dataset or a small-scale dataset.
Compared with the second-best method, MCoCo's ACC, NMI and Fscore are all improved by more than 10\% on Multi-COIL-20, Multi-Fashion, and Noisy-MNIST.
The main reason is that MCoCo makes different levels of spaces collaborate with each other while achieving their own consistency goals.
In this way, MCoCo can fully mine the multi-level consistent information of different views, which is utilized to guide the process of clustering.
For Noisy-MNIST, the second view has a chaotic clustering structure because of the white Gaussian noise. If multiple views are fused or mapped to the same space, this may cause the private information in the second view to have a negative impact on the final clustering effect.
Compared with AE$^2$-Nets, DCP, and DUA-Nets, which integrate multiple views into a unified representation for clustering, MCoCo can avoid the negative impact of private information of views with white Gaussian noise.
At the same time, it can be seen that compared with traditional clustering methods based on matrix decomposition, such as CMRL and LMSC, MCoCo has the advantage of complexity in dealing with the clustering of large-scale datasets.

\begin{figure}[t] 
\centering
\includegraphics[width=0.4\textwidth]{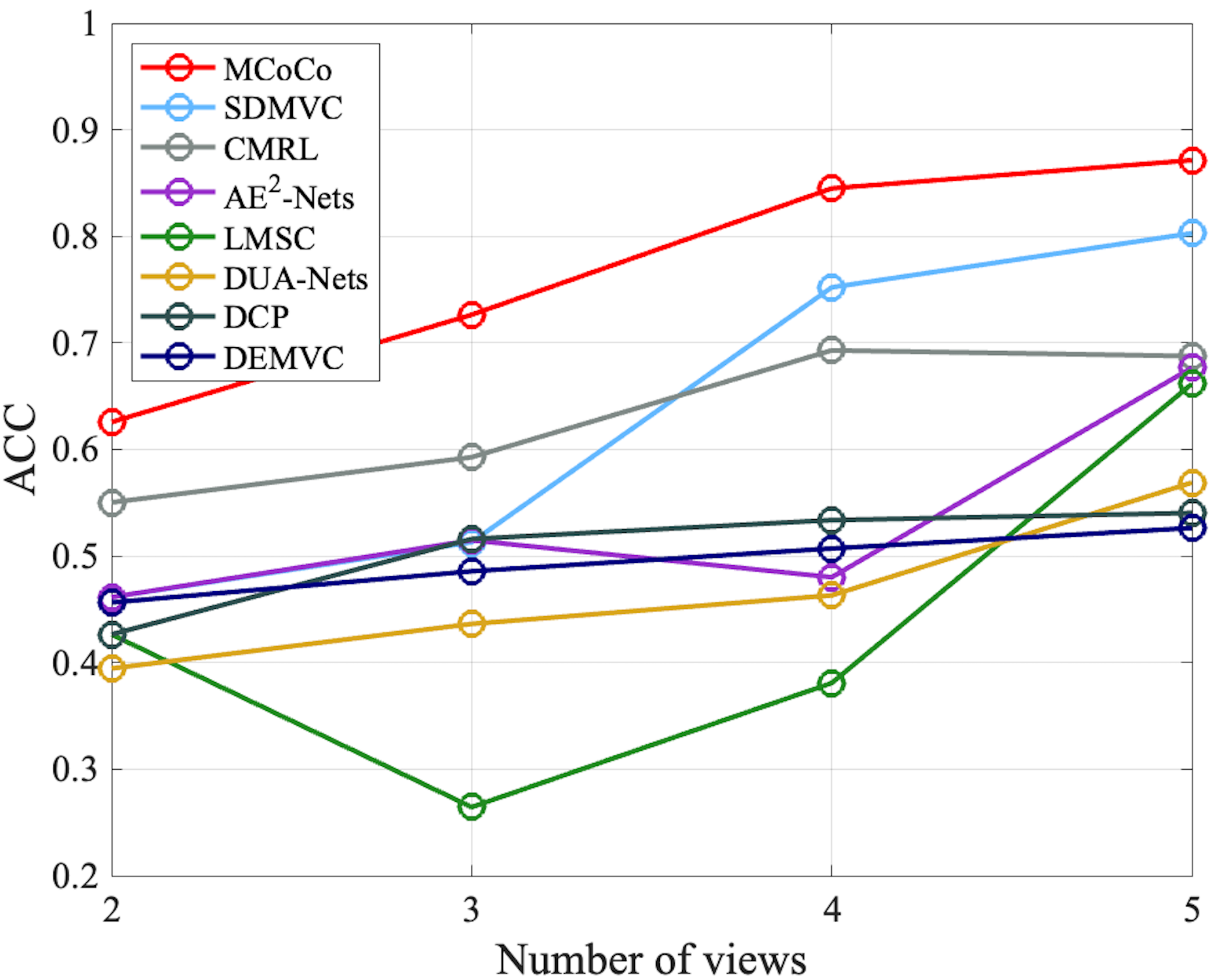}
\caption{
Clustering results on Caltech-$n$V.
}
\label{fg5} 
\end{figure}

In order to further verify our method, we conducted experiments on datasets with a variable number of views.
Figure \ref{fg5} shows the clustering results on Caltech-$n$V with different views.
We can see that the clustering performance of MCoCo grows steadily when the number of views increases.
At the same time, compared with AE$^2$-Nets, CMRL, and LMSC, which are methods that need to fuse different views, MCoCo can effectively avoid the negative impact of views with chaotic clustering structure on clustering results.
All these indicate that MCoCo can effectively mine different levels of consistent information, and the way of non-fusion can reduce the negative impact of unclear views on clustering.

\subsection{Visualization Result}
To vividly reveal the structure of the low-dimensional representation $\mathbf{Z}^{(m)}$, we visualize it achieved in Epoch 0, 10, 20, and 30 of the MCoCo learning process based on the t-SNE.
The visualization results are shown in Figure \ref{fg3}.
From Figure \ref{fg3}(d), we can see that the $\mathbf{Z}^{(m)}$ with a promising structure can be achieved by our method.
As can be seen from Figure \ref{fg3}(e), if the multi-level collaboration strategy is canceled, the clusters in the overlapping area can't be separated well.

To better demonstrate MCoCo's ability to separate clusters and obtain discriminative representations for each view on multiple datasets, we present in Figure \ref{fg4} the representations obtained by MCoCo for the first view of the remaining datasets in Table \ref{tb2} and Table \ref{tb3}.

\section{Model Analysis}

\subsection{Ablation Studies}
In order to verify the effectiveness of each part of our method, we conduct ablation studies here.
Consequently, we discuss the learning process of
our method with and without $\mathcal{L}_{Se}$ and $\mathcal{L}_{Ml}$.
Especially for canceling $\mathcal{L}_{Ml}$, we mean canceling the second half of Eq.\ref{ep11}, in other words, canceling multi-level collaboration.
We take the experiments on the BDGP dataset as an example.
The clustering results in metrics of ACC and NMI are reported in Table \ref{tb4}.
From the experimental results, we can find that: (1) In our proposed method, both $\mathcal{L}_{Se}$ and $\mathcal{L}_{Ml}$ can effectively improve the clustering performance;
(2) Compared with the original version, the ACC of the complete MCoCo can be improved by 30\%, which shows that multi-level collaboration is very important for clustering tasks;
(3) According to the clustering performance of MCoCo (view1) and MCoCo (view2), we can find that MCoCo can align different views well.

\begin{figure*}[t]
\centering
\subfloat[ACC in clustering task]{
\includegraphics[width=0.25\linewidth]{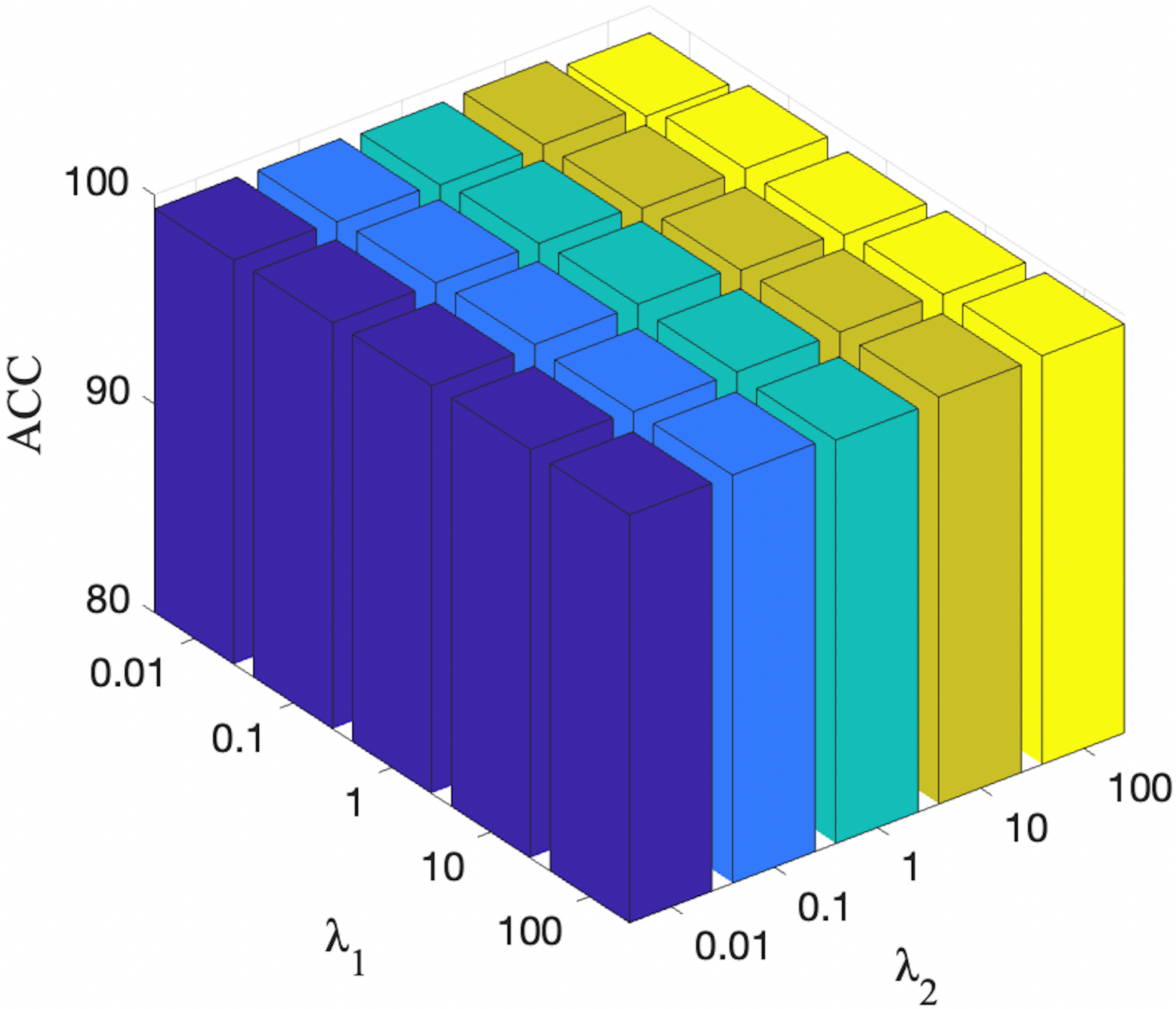}
}
\quad
\subfloat[NMI in clustering task]{
\includegraphics[width=0.25\linewidth]{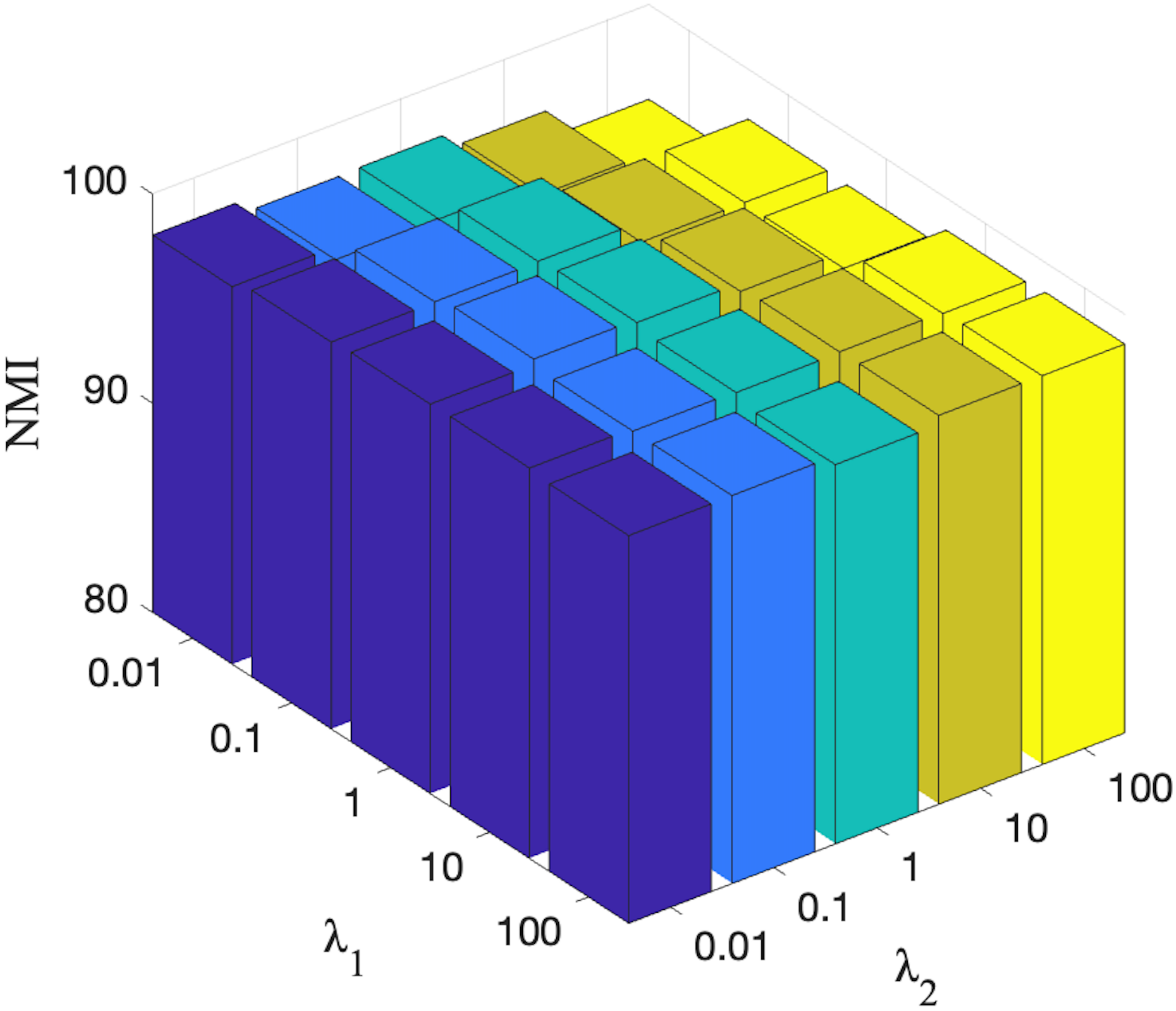}
}
\quad
\subfloat[Sensitivity of $\tau$]{
\includegraphics[width=0.28\linewidth]{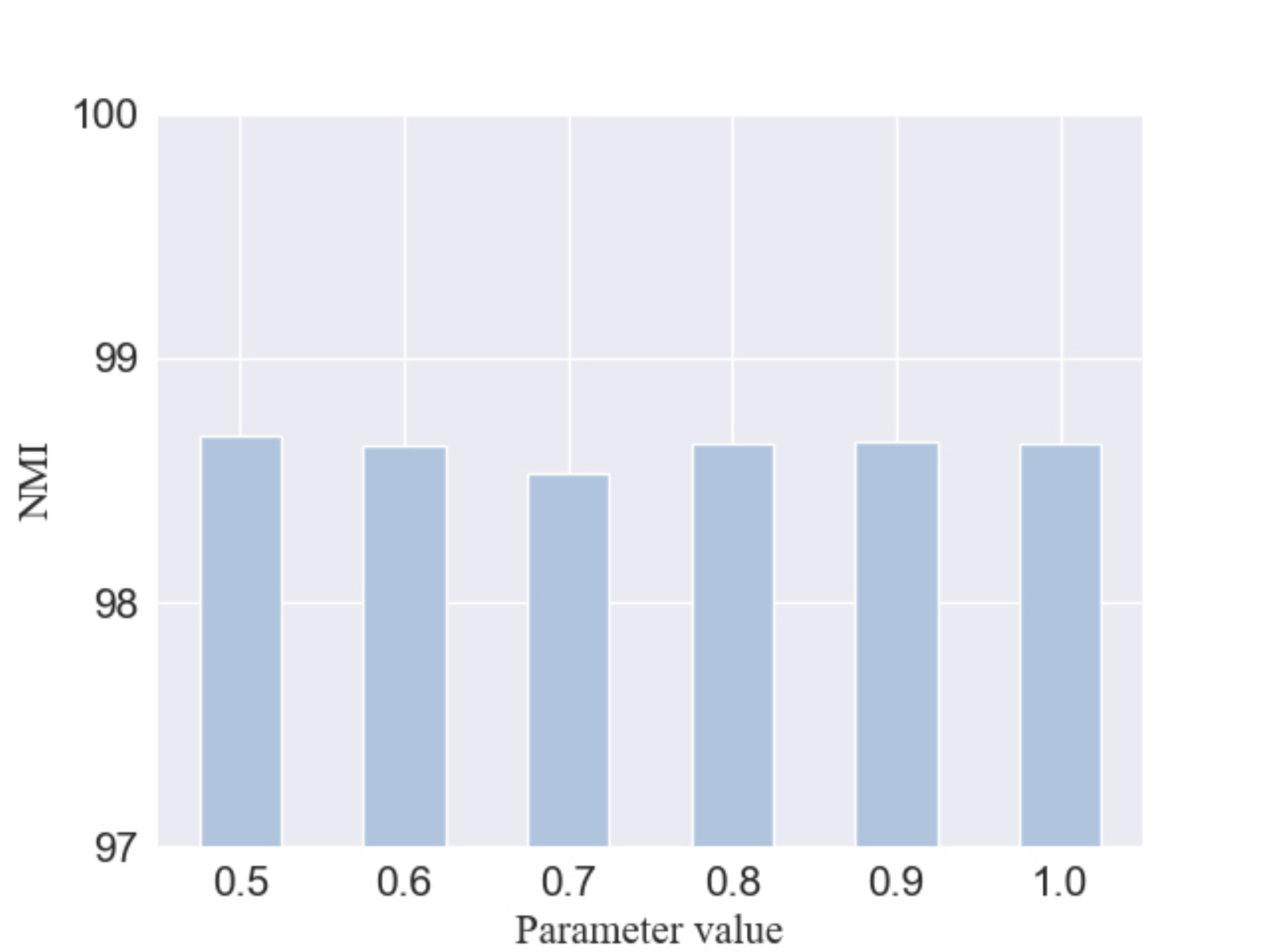}
}
\quad
\caption{Parameter sensitivity analysis of MCoCo on MNIST-USPS dataset.}
\label{fg6}
\end{figure*}
\begin{figure}[t]
\centering
\includegraphics[width=0.4\textwidth]{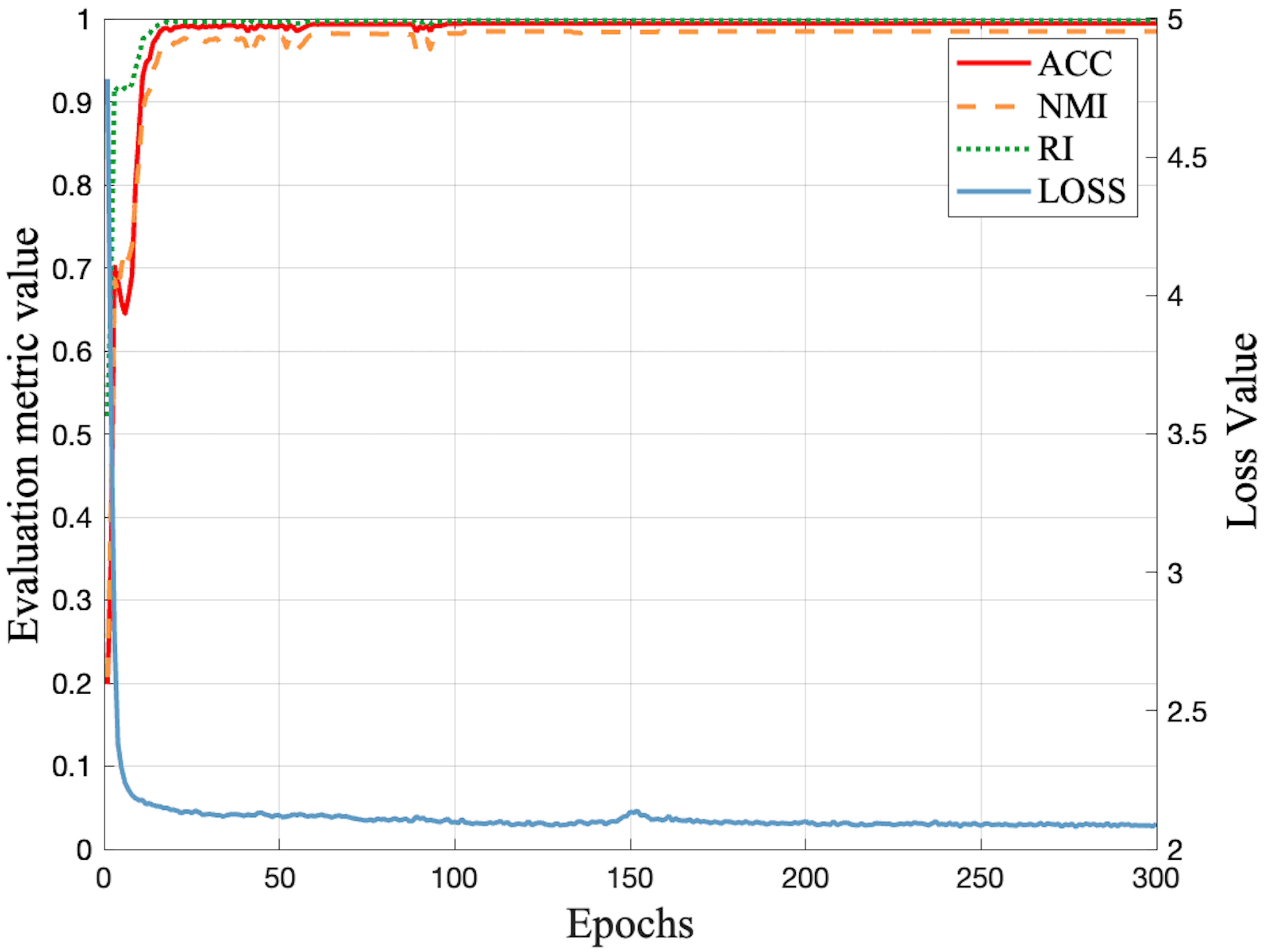} 
\caption{Convergence curve and clustering performance on MNIST-USPS dataset.
The X axis denotes training
epochs, and the left and right Y axis denote the evaluation metric value
and corresponding loss value, respectively.} 
\label{fg7}
\end{figure}
% \begin{figure*}[htbp]
%   \centering
%   \begin{minipage}[b]{0.3\textwidth}
%     \centering
%     \includegraphics[width=\textwidth]{Mcoco-acc.pdf}
%     \caption{ACC in clustering task}
%     \label{fig:sub1}
%   \end{minipage}
%   \hfill
%   \begin{minipage}[b]{0.3\textwidth}
%     \centering
%     \includegraphics[width=\textwidth]{Mcoco-nm.pdf}
%     \caption{NMI in clustering task}
%     \label{fig:sub2}
%   \end{minipage}
%   \hfill
%   \begin{minipage}[b]{0.3\textwidth}
%     \centering
%     \includegraphics[width=\textwidth]{mcoco-ss.pdf}
%     \caption{Sensitivity of $\tau$}
%     \label{fig:sub3}
%   \end{minipage}
%   \caption{Overall caption}
%   \label{fig:main}
% \end{figure*}

More specifically, it can be seen from Figure \ref{fg3}: According to Figure \ref{fg3}(a), it can be known that the different clusters in the first view of the BDGP dataset overlap seriously, and there are only two views in BDGP dataset.
It is difficult to collaboratively separate these overlapping clusters through another view.
If there is no collaboration of semantic labels in semantic space, the result will be as shown in Figure \ref{fg3}(e).
Different clusters in Figure \ref{fg3}(e) are difficult to separate, resulting in poor clustering performance.

\subsection{Parameter Sensitivity Analysis}
To explore the sensitivity of MCoCo to hyper-parameters, we first conducted an experiment on the MNIST-USPS dataset. In the experiment, we set different values to $\lambda _{1}$ and $\lambda _{2}$ in Eq.\ref{ep5} and explore their influence on the clustering task in the metric of ACC and NMI. 
In order to eliminate the randomness of the experiment and make the experimental results more reliable, our final results are all averaged by ten times clustering experiments.
The final experimental results are shown in  Figure \ref{fg6}(a) and Figure \ref{fg6}(b).
From the results, we can see that MCoCo is insensitive to the hyper-parameters $\lambda _{1}$ and $\lambda _{2}$.
In order to ensure the uniformity of all experiments, $\lambda _{1}$ and $\lambda _{2}$ are all fixed at 1 in other experiments in this paper.

For another hyper-parameter $\tau$ in Eq.\ref{ep7}, we set $\tau$ to $\{0.5, 0.6, 0.7, 0.8, 0.9, 1.0\}$ respectively, and do the same ten times for each clustering experiment to finally get the average value of metric NMI.
Figure \ref{fg6}(c) shows the experimental results on the dataset MNIST-USPS. The results show that MCoCo is also robust to $\tau$.
Actually, for all other experiments, we fixed $\tau$ at 0.5.

\subsection{Convergence Analysis}
To show the convergence properties of MCoCo, we take the experiment on the MNIST-USPS dataset and display the experimental results in Figure \ref{fg6}. It can be observed that the loss value drops rapidly in the first 15 epochs, with ACC, NMI, and RI continuously increasing. For other datasets, similar convergence properties can be achieved as well.

\begin{table}[t]
\centering  % 显示位置为中间
\caption{Ablation study on BDGP dataset.
In which (view 1) or (view 2) indicates that the final result is only obtained by cluster assignments of view1 or cluster assignments of view2.}  % 表格标题
\label{tb4}
\resizebox{0.45\textwidth}{!}{
\begin{tabular}{c|cc|cc}
\toprule
Method        & \multicolumn{2}{c|}{Components}                       & \multicolumn{2}{c}{Metrics} \\ \midrule
              & $\mathcal{L}_{Se}$                  & $\mathcal{L}_{Ml}$                  & ACC           & NMI          \\ \midrule
MCoCo(view 1) &                           &                           & 0.9848                   & 0.9484 \\
MCoCo(view 2) & \checkmark & \checkmark & 0.9800                   & 0.9479 \\
MCoCo         &                           &                           & \textbf{0.9872}                   & \textbf{0.9592} \\ \midrule
MCoCo(view 1) &                           &                           & 0.8120                   & 0.7821 \\
MCoCo(view 2) & \checkmark &                           & 0.8092                   & 0.7839 \\
MCoCo         &                           &                           & \textbf{0.8152}                   & \textbf{0.7984} \\ \midrule
MCoCo(view 1) &                           &                           & 0.6716                   & 0.6488 \\
MCoCo(view 2) &                           & \checkmark & 0.6696                   & 0.6478 \\
MCoCo         &                           &                           & \textbf{0.6840}                   & \textbf{0.6543} \\ \bottomrule
\end{tabular}}
\end{table}
\section{Conclusion}
In this paper, we propose a novel Multi-level Consistency Collaborative learning framework (MCoCo) for multi-view clustering, which can fully mine multi-level consistent information to guide the process of clustering.
While achieving consistency goals in different spaces, MCoCo can realize the collaboration between cluster assignments and consistent semantic labels.
Therefore, our method can get more discriminating clustering assignments for clustering. Meanwhile, MCoCo is also robust to some views with unclear clustering structure in a non-fusion manner.
Experimental results on several benchmark datasets verify the effectiveness of MCoCo over other state-of-the-art methods.

\section{Acknowledgment}
This work was supported by the National Key R\&D Pro-gram of China under Grant 2020AAA0109602.

\bibliographystyle{IEEEtran}
\bibliography{tcsvt}

\begin{thebibliography}{10}
\providecommand{\url}[1]{#1}
\csname url@samestyle\endcsname
\providecommand{\newblock}{\relax}
\providecommand{\bibinfo}[2]{#2}
\providecommand{\BIBentrySTDinterwordspacing}{\spaceskip=0pt\relax}
\providecommand{\BIBentryALTinterwordstretchfactor}{4}
\providecommand{\BIBentryALTinterwordspacing}{\spaceskip=\fontdimen2\font plus
\BIBentryALTinterwordstretchfactor\fontdimen3\font minus
  \fontdimen4\font\relax}
\providecommand{\BIBforeignlanguage}[2]{{%
\expandafter\ifx\csname l@#1\endcsname\relax
\typeout{** WARNING: IEEEtran.bst: No hyphenation pattern has been}%
\typeout{** loaded for the language `#1'. Using the pattern for}%
\typeout{** the default language instead.}%
\else
\language=\csname l@#1\endcsname
\fi
#2}}
\providecommand{\BIBdecl}{\relax}
\BIBdecl

\bibitem{geng2021uncertainty}
Y.~Geng, Z.~Han, C.~Zhang, and Q.~Hu, ``Uncertainty-aware multi-view
  representation learning,'' in \emph{Proceedings of the AAAI Conference on
  Artificial Intelligence}, vol.~35, no.~9, 2021, pp. 7545--7553.

\bibitem{zhang2018generalized}
C.~Zhang, H.~Fu, Q.~Hu, X.~Cao, Y.~Xie, D.~Tao, and D.~Xu, ``Generalized latent
  multi-view subspace clustering,'' \emph{IEEE transactions on pattern analysis
  and machine intelligence}, vol.~42, no.~1, pp. 86--99, 2018.

\bibitem{zhang2019ae2}
C.~Zhang, Y.~Liu, and H.~Fu, ``Ae2-nets: Autoencoder in autoencoder networks,''
  in \emph{Proceedings of the IEEE/CVF conference on computer vision and
  pattern recognition}, 2019, pp. 2577--2585.

\bibitem{zheng2020feature}
Q.~Zheng, J.~Zhu, Z.~Li, S.~Pang, J.~Wang, and Y.~Li, ``Feature concatenation
  multi-view subspace clustering,'' \emph{Neurocomputing}, vol. 379, pp.
  89--102, 2020.

\bibitem{jia2021multi}
Y.~Jia, H.~Liu, J.~Hou, S.~Kwong, and Q.~Zhang, ``Multi-view spectral
  clustering tailored tensor low-rank representation,'' \emph{IEEE Transactions
  on Circuits and Systems for Video Technology}, vol.~31, no.~12, pp.
  4784--4797, 2021.

\bibitem{yin2018multiview}
M.~Yin, J.~Gao, S.~Xie, and Y.~Guo, ``Multiview subspace clustering via
  tensorial t-product representation,'' \emph{IEEE transactions on neural
  networks and learning systems}, vol.~30, no.~3, pp. 851--864, 2018.

\bibitem{xie2020robust}
Y.~Xie, J.~Liu, Y.~Qu, D.~Tao, W.~Zhang, L.~Dai, and L.~Ma, ``Robust kernelized
  multiview self-representation for subspace clustering,'' \emph{IEEE
  transactions on neural networks and learning systems}, vol.~32, no.~2, pp.
  868--881, 2020.

\bibitem{guo2021rank}
J.~Guo, Y.~Sun, J.~Gao, Y.~Hu, and B.~Yin, ``Rank consistency induced multiview
  subspace clustering via low-rank matrix factorization,'' \emph{IEEE
  Transactions on Neural Networks and Learning Systems}, 2021.

\bibitem{kang2020large}
Z.~Kang, W.~Zhou, Z.~Zhao, J.~Shao, M.~Han, and Z.~Xu, ``Large-scale multi-view
  subspace clustering in linear time,'' in \emph{Proceedings of the AAAI
  conference on artificial intelligence}, vol.~34, no.~04, 2020, pp.
  4412--4419.

\bibitem{chen2021low}
Y.~Chen, X.~Xiao, C.~Peng, G.~Lu, and Y.~Zhou, ``Low-rank tensor graph learning
  for multi-view subspace clustering,'' \emph{IEEE Transactions on Circuits and
  Systems for Video Technology}, vol.~32, no.~1, pp. 92--104, 2021.

\bibitem{lan2021generalized}
M.~Lan, M.~Meng, J.~Yu, and J.~Wu, ``Generalized multi-view collaborative
  subspace clustering,'' \emph{IEEE Transactions on Circuits and Systems for
  Video Technology}, vol.~32, no.~6, pp. 3561--3574, 2021.

\bibitem{liu2013multi}
J.~Liu, C.~Wang, J.~Gao, and J.~Han, ``Multi-view clustering via joint
  nonnegative matrix factorization,'' in \emph{Proceedings of the 2013 SIAM
  international conference on data mining}.\hskip 1em plus 0.5em minus
  0.4em\relax SIAM, 2013, pp. 252--260.

\bibitem{wang2018multiview}
Y.~Wang, L.~Wu, X.~Lin, and J.~Gao, ``Multiview spectral clustering via
  structured low-rank matrix factorization,'' \emph{IEEE transactions on neural
  networks and learning systems}, vol.~29, no.~10, pp. 4833--4843, 2018.

\bibitem{yang2020uniform}
Z.~Yang, N.~Liang, W.~Yan, Z.~Li, and S.~Xie, ``Uniform distribution
  non-negative matrix factorization for multiview clustering,'' \emph{IEEE
  transactions on cybernetics}, vol.~51, no.~6, pp. 3249--3262, 2020.

\bibitem{nie2017self}
F.~Nie, J.~Li, X.~Li \emph{et~al.}, ``Self-weighted multiview clustering with
  multiple graphs.'' in \emph{IJCAI}, 2017, pp. 2564--2570.

\bibitem{wang2019gmc}
H.~Wang, Y.~Yang, and B.~Liu, ``Gmc: Graph-based multi-view clustering,''
  \emph{IEEE Transactions on Knowledge and Data Engineering}, vol.~32, no.~6,
  pp. 1116--1129, 2019.

\bibitem{peng2019comic}
X.~Peng, Z.~Huang, J.~Lv, H.~Zhu, and J.~T. Zhou, ``Comic: Multi-view
  clustering without parameter selection,'' in \emph{International conference
  on machine learning}.\hskip 1em plus 0.5em minus 0.4em\relax PMLR, 2019, pp.
  5092--5101.

\bibitem{zheng2022graph}
Q.~Zheng, J.~Zhu, Z.~Li, and H.~Tang, ``Graph-guided unsupervised multiview
  representation learning,'' \emph{IEEE Transactions on Circuits and Systems
  for Video Technology}, vol.~33, no.~1, pp. 146--159, 2022.

\bibitem{wong2019clustering}
W.~K. Wong, N.~Han, X.~Fang, S.~Zhan, and J.~Wen, ``Clustering
  structure-induced robust multi-view graph recovery,'' \emph{IEEE Transactions
  on Circuits and Systems for Video Technology}, vol.~30, no.~10, pp.
  3584--3597, 2019.

\bibitem{wang2022towards}
H.~Wang, G.~Jiang, J.~Peng, R.~Deng, and X.~Fu, ``Towards adaptive consensus
  graph: Multi-view clustering via graph collaboration,'' \emph{IEEE
  Transactions on Multimedia}, 2022.

\bibitem{mildenhall2021nerf}
B.~Mildenhall, P.~P. Srinivasan, M.~Tancik, J.~T. Barron, R.~Ramamoorthi, and
  R.~Ng, ``Nerf: Representing scenes as neural radiance fields for view
  synthesis,'' \emph{Communications of the ACM}, vol.~65, no.~1, pp. 99--106,
  2021.

\bibitem{xu2019learning}
H.~Xu, P.~Liang, W.~Yu, J.~Jiang, and J.~Ma, ``Learning a generative model for
  fusing infrared and visible images via conditional generative adversarial
  network with dual discriminators.'' in \emph{IJCAI}, 2019, pp. 3954--3960.

\bibitem{tao2017deep}
D.~Tao, Y.~Guo, B.~Yu, J.~Pang, and Z.~Yu, ``Deep multi-view feature learning
  for person re-identification,'' \emph{IEEE Transactions on Circuits and
  Systems for Video Technology}, vol.~28, no.~10, pp. 2657--2666, 2017.

\bibitem{li2019deep}
Z.~Li, Q.~Wang, Z.~Tao, Q.~Gao, Z.~Yang \emph{et~al.}, ``Deep adversarial
  multi-view clustering network.'' in \emph{IJCAI}, 2019, pp. 2952--2958.

\bibitem{xu2021multi}
J.~Xu, Y.~Ren, H.~Tang, X.~Pu, X.~Zhu, M.~Zeng, and L.~He, ``Multi-vae:
  Learning disentangled view-common and view-peculiar visual representations
  for multi-view clustering,'' in \emph{Proceedings of the IEEE/CVF
  International Conference on Computer Vision}, 2021, pp. 9234--9243.

\bibitem{xu2022multi}
J.~Xu, H.~Tang, Y.~Ren, L.~Peng, X.~Zhu, and L.~He, ``Multi-level feature
  learning for contrastive multi-view clustering,'' in \emph{Proceedings of the
  IEEE/CVF Conference on Computer Vision and Pattern Recognition}, 2022, pp.
  16\,051--16\,060.

\bibitem{lin2021completer}
Y.~Lin, Y.~Gou, Z.~Liu, B.~Li, J.~Lv, and X.~Peng, ``Completer: Incomplete
  multi-view clustering via contrastive prediction,'' in \emph{Proceedings of
  the IEEE/CVF Conference on Computer Vision and Pattern Recognition}, 2021,
  pp. 11\,174--11\,183.

\bibitem{zheng2021collaborative}
Q.~Zheng, J.~Zhu, and Z.~Li, ``Collaborative unsupervised multi-view
  representation learning,'' \emph{IEEE Transactions on Circuits and Systems
  for Video Technology}, 2021.

\bibitem{xu2022self}
J.~Xu, Y.~Ren, H.~Tang, Z.~Yang, L.~Pan, Y.~Yang, X.~Pu, S.~Y. Philip, and
  L.~He, ``Self-supervised discriminative feature learning for deep multi-view
  clustering,'' \emph{IEEE Transactions on Knowledge and Data Engineering},
  2022.

\bibitem{xu2021deep}
J.~Xu, Y.~Ren, G.~Li, L.~Pan, C.~Zhu, and Z.~Xu, ``Deep embedded multi-view
  clustering with collaborative training,'' \emph{Information Sciences}, vol.
  573, pp. 279--290, 2021.

\bibitem{xie2016unsupervised}
J.~Xie, R.~Girshick, and A.~Farhadi, ``Unsupervised deep embedding for
  clustering analysis,'' in \emph{International conference on machine
  learning}.\hskip 1em plus 0.5em minus 0.4em\relax PMLR, 2016, pp. 478--487.

\bibitem{cheng2021multi}
J.~Cheng, Q.~Wang, Z.~Tao, D.~Xie, and Q.~Gao, ``Multi-view attribute graph
  convolution networks for clustering,'' in \emph{Proceedings of the
  Twenty-Ninth International Conference on International Joint Conferences on
  Artificial Intelligence}, 2021, pp. 2973--2979.

\bibitem{guo2017improved}
X.~Guo, L.~Gao, X.~Liu, and J.~Yin, ``Improved deep embedded clustering with
  local structure preservation.'' in \emph{Ijcai}, 2017, pp. 1753--1759.

\bibitem{ghasedi2017deep}
K.~Ghasedi~Dizaji, A.~Herandi, C.~Deng, W.~Cai, and H.~Huang, ``Deep clustering
  via joint convolutional autoencoder embedding and relative entropy
  minimization,'' in \emph{Proceedings of the IEEE international conference on
  computer vision}, 2017, pp. 5736--5745.

\bibitem{mukherjee2019clustergan}
S.~Mukherjee, H.~Asnani, E.~Lin, and S.~Kannan, ``Clustergan: Latent space
  clustering in generative adversarial networks,'' in \emph{Proceedings of the
  AAAI conference on artificial intelligence}, vol.~33, no.~01, 2019, pp.
  4610--4617.

\bibitem{zhao2017multi}
J.~Zhao, X.~Xie, X.~Xu, and S.~Sun, ``Multi-view learning overview: Recent
  progress and new challenges,'' \emph{Information Fusion}, vol.~38, pp.
  43--54, 2017.

\bibitem{xu2013survey}
C.~Xu, D.~Tao, and C.~Xu, ``A survey on multi-view learning,'' \emph{arXiv
  preprint arXiv:1304.5634}, 2013.

\bibitem{wang2015deep}
W.~Wang, R.~Arora, K.~Livescu, and J.~Bilmes, ``On deep multi-view
  representation learning,'' in \emph{International conference on machine
  learning}.\hskip 1em plus 0.5em minus 0.4em\relax PMLR, 2015, pp. 1083--1092.

\bibitem{hotelling1992relations}
H.~Hotelling, ``Relations between two sets of variates,'' \emph{Breakthroughs
  in statistics: methodology and distribution}, pp. 162--190, 1992.

\bibitem{akaho2006kernel}
S.~Akaho, ``A kernel method for canonical correlation analysis,'' \emph{arXiv
  preprint cs/0609071}, 2006.

\bibitem{andrew2013deep}
G.~Andrew, R.~Arora, J.~Bilmes, and K.~Livescu, ``Deep canonical correlation
  analysis,'' in \emph{International conference on machine learning}.\hskip 1em
  plus 0.5em minus 0.4em\relax PMLR, 2013, pp. 1247--1255.

\bibitem{shao2016deep}
J.~Shao, L.~Wang, Z.~Zhao, A.~Cai \emph{et~al.}, ``Deep canonical correlation
  analysis with progressive and hypergraph learning for cross-modal
  retrieval,'' \emph{Neurocomputing}, vol. 214, pp. 618--628, 2016.

\bibitem{shen2015unified}
X.~Shen, Q.~Sun, and Y.~Yuan, ``A unified multiset canonical correlation
  analysis framework based on graph embedding for multiple feature
  extraction,'' \emph{Neurocomputing}, vol. 148, pp. 397--408, 2015.

\bibitem{chen2018canonical}
J.~Chen, G.~Wang, Y.~Shen, and G.~B. Giannakis, ``Canonical correlation
  analysis of datasets with a common source graph,'' \emph{IEEE Transactions on
  Signal Processing}, vol.~66, no.~16, pp. 4398--4408, 2018.

\bibitem{zheng2023comprehensive}
Q.~Zheng, J.~Zhu, Z.~Li, Z.~Tian, and C.~Li, ``Comprehensive multi-view
  representation learning,'' \emph{Information Fusion}, vol.~89, pp. 198--209,
  2023.

\bibitem{huang2021deep}
Z.~Huang, J.~T. Zhou, H.~Zhu, C.~Zhang, J.~Lv, and X.~Peng, ``Deep spectral
  representation learning from multi-view data,'' \emph{IEEE Transactions on
  Image Processing}, vol.~30, pp. 5352--5362, 2021.

\bibitem{wan2021multill}
Z.~Wan, C.~Zhang, P.~Zhu, and Q.~Hu, ``Multi-view information-bottleneck
  representation learning,'' in \emph{Proceedings of the AAAI Conference on
  Artificial Intelligence}, vol.~35, no.~11, 2021, pp. 10\,085--10\,092.

\bibitem{hu2021akm}
Y.~Hu, Z.~Song, B.~Wang, J.~Gao, Y.~Sun, and B.~Yin, ``Akm 3 c: Adaptive
  k-multiple-means for multi-view clustering,'' \emph{IEEE Transactions on
  Circuits and Systems for Video Technology}, vol.~31, no.~11, pp. 4214--4226,
  2021.

\bibitem{chen2020simple}
T.~Chen, S.~Kornblith, M.~Norouzi, and G.~Hinton, ``A simple framework for
  contrastive learning of visual representations,'' in \emph{International
  conference on machine learning}.\hskip 1em plus 0.5em minus 0.4em\relax PMLR,
  2020, pp. 1597--1607.

\bibitem{xie2021detco}
E.~Xie, J.~Ding, W.~Wang, X.~Zhan, H.~Xu, P.~Sun, Z.~Li, and P.~Luo, ``Detco:
  Unsupervised contrastive learning for object detection,'' in
  \emph{Proceedings of the IEEE/CVF International Conference on Computer
  Vision}, 2021, pp. 8392--8401.

\bibitem{niu2022spice}
C.~Niu, H.~Shan, and G.~Wang, ``Spice: Semantic pseudo-labeling for image
  clustering,'' \emph{IEEE Transactions on Image Processing}, vol.~31, pp.
  7264--7278, 2022.

\bibitem{van2020scan}
W.~Van~Gansbeke, S.~Vandenhende, S.~Georgoulis, M.~Proesmans, and L.~Van~Gool,
  ``Scan: Learning to classify images without labels,'' in \emph{Computer
  Vision--ECCV 2020: 16th European Conference, Glasgow, UK, August 23--28,
  2020, Proceedings, Part X}.\hskip 1em plus 0.5em minus 0.4em\relax Springer,
  2020, pp. 268--285.

\bibitem{li2021contrastive}
Y.~Li, P.~Hu, Z.~Liu, D.~Peng, J.~T. Zhou, and X.~Peng, ``Contrastive
  clustering,'' in \emph{Proceedings of the AAAI Conference on Artificial
  Intelligence}, vol.~35, no.~10, 2021, pp. 8547--8555.

\bibitem{roy2021self}
S.~Roy and A.~Etemad, ``Self-supervised contrastive learning of multi-view
  facial expressions,'' in \emph{Proceedings of the 2021 International
  Conference on Multimodal Interaction}, 2021, pp. 253--257.

\bibitem{lin2022contrastive}
F.~Lin, B.~Bai, K.~Bai, Y.~Ren, P.~Zhao, and Z.~Xu, ``Contrastive multi-view
  hyperbolic hierarchical clustering,'' \emph{arXiv preprint arXiv:2205.02618},
  2022.

\bibitem{hassani2020contrastive}
K.~Hassani and A.~H. Khasahmadi, ``Contrastive multi-view representation
  learning on graphs,'' in \emph{International conference on machine
  learning}.\hskip 1em plus 0.5em minus 0.4em\relax PMLR, 2020, pp. 4116--4126.

\bibitem{wan2021multi}
Z.~Wan, C.~Zhang, P.~Zhu, and Q.~Hu, ``Multi-view information-bottleneck
  representation learning,'' in \emph{Proceedings of the AAAI Conference on
  Artificial Intelligence}, vol.~35, no.~11, 2021, pp. 10\,085--10\,092.

\bibitem{cai2012joint}
X.~Cai, H.~Wang, H.~Huang, and C.~Ding, ``Joint stage recognition and
  anatomical annotation of drosophila gene expression patterns,''
  \emph{Bioinformatics}, vol.~28, no.~12, pp. i16--i24, 2012.

\bibitem{xiao2017fashion}
H.~Xiao, K.~Rasul, and R.~Vollgraf, ``Fashion-mnist: a novel image dataset for
  benchmarking machine learning algorithms,'' \emph{arXiv preprint
  arXiv:1708.07747}, 2017.

\bibitem{fei2004learning}
L.~Fei-Fei, R.~Fergus, and P.~Perona, ``Learning generative visual models from
  few training examples: An incremental bayesian approach tested on 101 object
  categories,'' in \emph{2004 conference on computer vision and pattern
  recognition workshop}.\hskip 1em plus 0.5em minus 0.4em\relax IEEE, 2004, pp.
  178--178.

\bibitem{zhang2017latent}
C.~Zhang, Q.~Hu, H.~Fu, P.~Zhu, and X.~Cao, ``Latent multi-view subspace
  clustering,'' in \emph{Proceedings of the IEEE conference on computer vision
  and pattern recognition}, 2017, pp. 4279--4287.

\bibitem{lin2022dual}
Y.~Lin, Y.~Gou, X.~Liu, J.~Bai, J.~Lv, and X.~Peng, ``Dual contrastive
  prediction for incomplete multi-view representation learning,'' \emph{IEEE
  Transactions on Pattern Analysis and Machine Intelligence}, 2022.

\bibitem{glorot2011deep}
X.~Glorot, A.~Bordes, and Y.~Bengio, ``Deep sparse rectifier neural networks,''
  in \emph{Proceedings of the fourteenth international conference on artificial
  intelligence and statistics}.\hskip 1em plus 0.5em minus 0.4em\relax JMLR
  Workshop and Conference Proceedings, 2011, pp. 315--323.

\bibitem{kingma2014adam}
D.~P. Kingma and J.~Ba, ``Adam: A method for stochastic optimization,''
  \emph{arXiv preprint arXiv:1412.6980}, 2014.

\bibitem{paszke2019pytorch}
A.~Paszke, S.~Gross, F.~Massa, A.~Lerer, J.~Bradbury, G.~Chanan, T.~Killeen,
  Z.~Lin, N.~Gimelshein, L.~Antiga \emph{et~al.}, ``Pytorch: An imperative
  style, high-performance deep learning library,'' \emph{Advances in neural
  information processing systems}, vol.~32, 2019.

\end{thebibliography}
% \bibitem{ref1}
% {\it{Mathematics Into Type}}. American Mathematical Society. [Online]. Available: https://www.ams.org/arc/styleguide/mit-2.pdf

% \bibitem{ref2}
% T. W. Chaundy, P. R. Barrett and C. Batey, {\it{The Printing of Mathematics}}. London, U.K., Oxford Univ. Press, 1954.

% \bibitem{ref3}
% F. Mittelbach and M. Goossens, {\it{The \LaTeX Companion}}, 2nd ed. Boston, MA, USA: Pearson, 2004.

% \bibitem{ref4}
% G. Gr\"atzer, {\it{More Math Into LaTeX}}, New York, NY, USA: Springer, 2007.

% \bibitem{ref5}M. Letourneau and J. W. Sharp, {\it{AMS-StyleGuide-online.pdf,}} American Mathematical Society, Providence, RI, USA, [Online]. Available: http://www.ams.org/arc/styleguide/index.html

% \bibitem{ref6}
% H. Sira-Ramirez, ``On the sliding mode control of nonlinear systems,'' \textit{Syst. Control Lett.}, vol. 19, pp. 303--312, 1992.

% \bibitem{ref7}
% A. Levant, ``Exact differentiation of signals with unbounded higher derivatives,''  in \textit{Proc. 45th IEEE Conf. Decis.
% Control}, San Diego, CA, USA, 2006, pp. 5585--5590. DOI: 10.1109/CDC.2006.377165.

% \bibitem{ref8}
% M. Fliess, C. Join, and H. Sira-Ramirez, ``Non-linear estimation is easy,'' \textit{Int. J. Model., Ident. Control}, vol. 4, no. 1, pp. 12--27, 2008.

% \bibitem{ref9}
% R. Ortega, A. Astolfi, G. Bastin, and H. Rodriguez, ``Stabilization of food-chain systems using a port-controlled Hamiltonian description,'' in \textit{Proc. Amer. Control Conf.}, Chicago, IL, USA,
% 2000, pp. 2245--2249.

% \end{thebibliography}

\newpage

\vfill

\end{document}